\documentclass{article}

\usepackage{arxiv}

\usepackage[utf8]{inputenc} 
\usepackage[T1]{fontenc}   
\usepackage{hyperref}   
\usepackage{url}      
\usepackage{booktabs}
\usepackage{amsfonts} 
\usepackage{nicefrac} 
\usepackage{microtype}  
\usepackage{lipsum}	
\usepackage{graphicx}
\usepackage{natbib}
\usepackage{doi}

\usepackage{multirow}
\usepackage{listings}
\usepackage{amsmath}
\usepackage{amssymb}

\usepackage{algorithm}
\usepackage{algpseudocode}
\usepackage{enumitem}
\usepackage{xspace}
\usepackage{caption}
\usepackage{comment}
\usepackage{subcaption}
\usepackage[export]{adjustbox}

\newcommand{\R}{\mathbb{R}}
\def\ours{\textsc{HashEvict}\xspace}

\title{HashEvict: A Pre-Attention KV Cache Eviction Strategy using Locality-Sensitive Hashing}
	
\author{Minghui Liu \thanks{Equal contributions.}\\
	University of Maryland\\
	\texttt{minghui@umd.edu} \\
	\And
	Tahseen Rabbani \footnotemark[1] \\
	Yale University\\
	\texttt{tahseen.rabbani@yale.edu} \\
	\AND
	Tony O'Halloran \\
	University of Galway \\
	\texttt{t.ohalloran8@universityofgalway.ie} \\
	\And
	Ananth Sankaralingam \\
	University of Maryland \\
	\texttt{ananths1@terpmail.umd.edu} \\
    \And
    Mary-Anne Hartley \\
    Yale University\\
	\texttt{mary-anne.hartley@yale.edu} \\
    \And
	Furong Huang \\
	University of Maryland \\
    Capital One \\
	\texttt{furongh@cs.umd.edu} \\
    \And 
    Cornelia Fermüller \\ 
    University of Maryland\\
	\texttt{fermulcm@umd.edu} \\
    \And
    Yiannis Aloimonos \\
    University of Maryland\\
	\texttt{jyaloimo@umd.edu}
}

\date{}

\renewcommand{\shorttitle}{\textit{arXiv} Template}

\hypersetup{
pdftitle={HashEvict: },
pdfsubject={artificial intelligence, machine learning, natural language processing},
pdfauthor={Minghui Liu, Tahseen Rabbani, Tony O'Halloran, Anath Sankaralingam, Mary-Anne Hartley, Furong Huang, Cornelia Fermüller, Yiannis Aloimonos},
pdfkeywords={KV Cache Management, Locality-sensitive Hashing, Efficient LLM Inference},
}

\begin{document}
\maketitle

\begin{abstract}
The key-value (KV) cache in large language models (LLMs) markedly accelerates attention and inference. However, since it requires storing the key and value embeddings of past tokens it can quickly consume a significant amount of GPU memory. In this work, we introduce \ours, a KV cache compression strategy that uses locality-sensitive hashing (LSH) to efficiently estimate attention and evict unimportant tokens. At every decoding step, the key and value of the current token replace the embeddings of a token estimated to produce the lowest attention score. We project key and query embeddings to dimensionally-reduced binary hash codes and locate low-attention tokens via a lightweight Hamming distance calculation, all of which is efficiently computable on the GPU. Unlike existing compression strategies that compute attention to determine token retention, \ours makes these decisions pre-attention, thereby reducing computational costs. We demonstrate that \ours can compress the KV cache by 30\%-70\% while maintaining high performance across reasoning, multiple-choice, long-context retrieval and summarization tasks. Our method is fast: we achieve 1.5-2x prefill speed and competitive decoding speed against baseline methods such as H$_2$O, Scissorhands, and 17$\times$ prefill/2x decoding speed against FastGen. 
\end{abstract}

\keywords{KV cache \and locality-sensitive hashing \and efficient LLM inference \and token eviction}

\section{Introduction}
The advent of large language models (LLMs) has enabled sharp improvements over innumerable downstream natural language processing (NLP) tasks, such as summarization and dialogue generation \citep{zhao2023survey, wei2022emergent}. The hallmark feature of LLMs, the attention module \citep{bahdanau2014neural,luong2015effective, vaswani2017attention}, enables contextual processing over sequences of tokens. To avoid repeated dot products over key and value embeddings of tokens, a key-value (KV) cache is maintained in VRAM to maintain these calculations. This technique is particularly popular with decoder LLMs. 

However, the size of the KV cache scales quadratically with sequence length $n$ and linearly with the number of attention layers and heads. Assuming the size of the KV cache is $n$ tokens, for each new decoded token, $n$ attention scores need to be added which requires a total of $\mathcal{O}(dn^2)$ computation, where $d$ is the projection dimension, and $\mathcal{O}(n^2)$ storage. For example, maintaining the KV cache for a sequence of 4K tokens in half-precision (FP16) can require approximately $\sim$16GB of memory for most models within the Llama 3 family \citep{dubey2024llama}. These memory costs are exacerbated with batched inference and result in high decoding latency \citep{fu2024challenges}. Consequently, there is significant interest in compressing the size of the KV cache to enable longer context windows and low-resource, on-device deployment.

An emerging strategy for reducing the size of the KV cache is \textit{token eviction}. This approach drops the key and value embeddings for past tokens in the cache, skipping future attention calculations involving these tokens. 

Various token eviction/retention policies have been explored in recent literature, including

the profiling of token type preferences \citep{ge2023model}, retention of heavy-hitter tokens \citep{zhang2024h2o, zhang2024q}, and dropping tokens based on the high $L_2$ norms of their key embeddings \citep{devoto2024simple}.

The latter approach \citep{devoto2024simple} is intriguing as eviction decisions are performed pre-attention. 
However, this $L_2$ dropout strategy in inclined towards long-context retrieval tasks. It developed based on an empirical observation that smaller norm of key embedding correlates with higher attention score. For long-context retrieval tasks, high-attention score tokens are the most important tokens since the question's text will overlap with the piece of context that needs to be retrieved. Thus, it is specialized to retain only those tokens with the highest attention, which we find unsuitable for free response reasoning tasks.  Existing literature suggests that retaining tokens with a diverse spectrum of attention scores (skewing high) is necessary \citep{guo2024attention, zhang2024h2o,long2023beyond}.

\begin{figure}[ht!]
    \begin{subfigure}[t!]{0.55\textwidth}
        \centering
        \includegraphics[width=\textwidth]{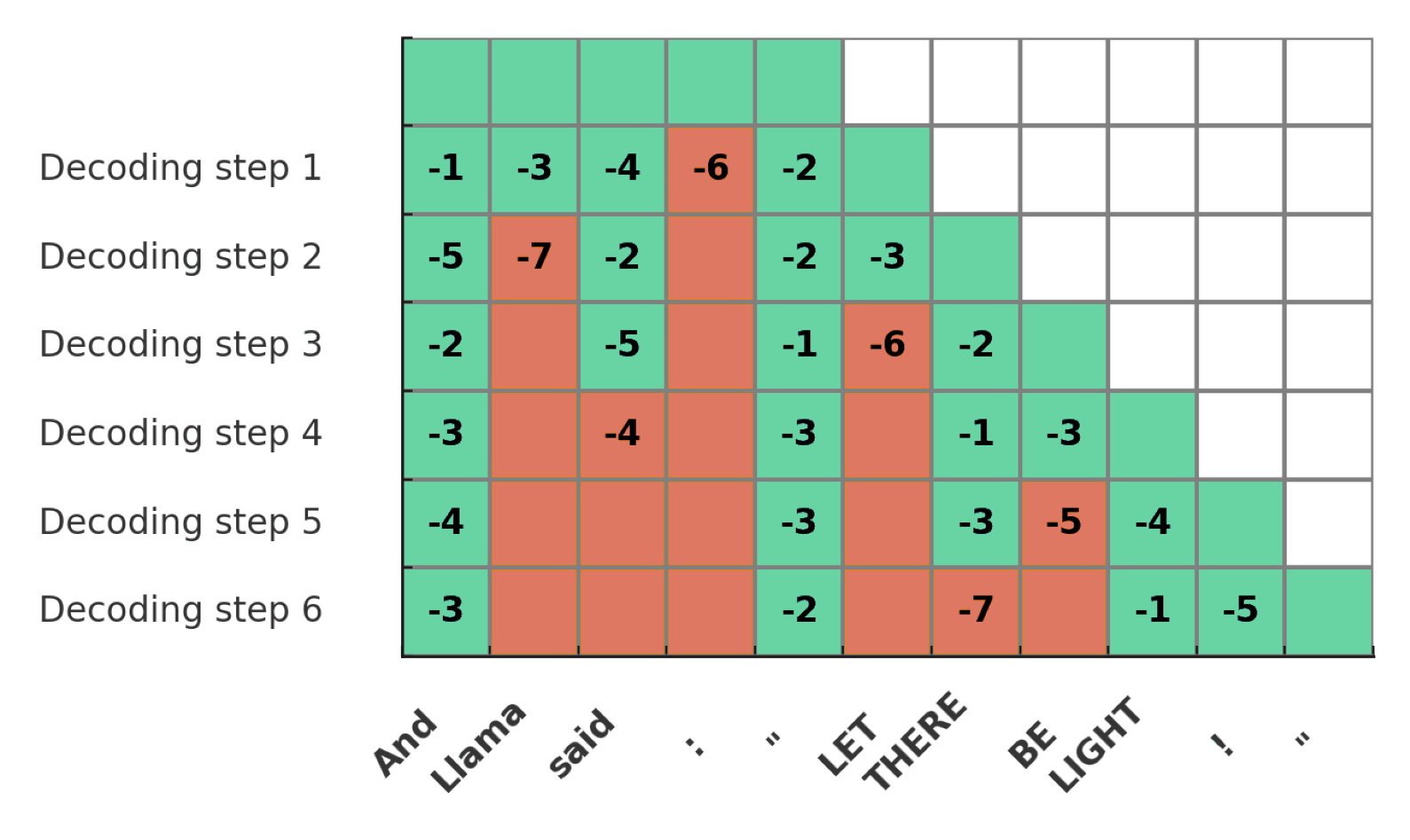}  
        \caption{KV cache during decoding}
        \label{fig:full-strategy}
    \end{subfigure}
    \hfill
    \begin{subfigure}[t!]{0.43\textwidth}
        \centering
        \includegraphics[trim = {0 -4.73cm -1.7cm 0 -10cm}, width=\textwidth]{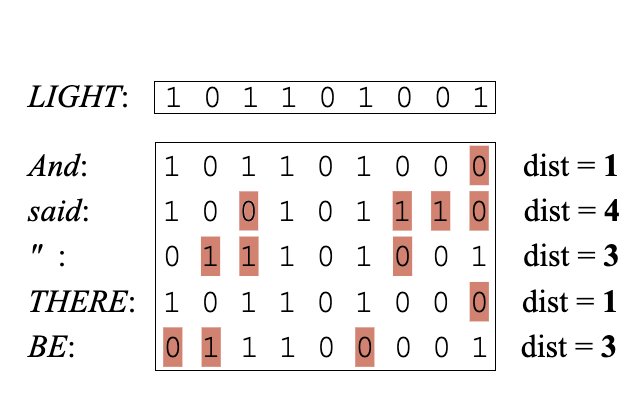}
        \caption{LSH comparison at decoding step 4}
        \label{fig:decoding-step}
    \end{subfigure}
    \caption{\textbf{An abstract visualization of \ours eviction strategy.} Figure \ref{fig:full-strategy} depicts the strategy for several decoding steps. The cache can only maintain 5 tokens due to memory constraints. At each decoding step, \ours projects the query embedding of the current token $i$ and all previous key embeddings to \textit{binary hash \underline{codes}}. \ours then measures the negative of Hamming distances between the query \textit{\underline{code}} of token $i$ and key \textit{\underline{codes}} of all tokens $j$ in the cache. Each step, \ours evicts the key/values of the token with the lowest score (marked as red) from the cache. Figure \ref{fig:decoding-step} depicts the LSH comparison for decoding step 4, marking the token \textit{``said"} for removal, as its high Hamming indicates low cosine similarity (and thus, low attention). }
    \label{visualization-lshe}
\end{figure}

\textit{Is there a non-attentive KV cache compression strategy that is performant over a wide variety of tasks, including multiple-choice, summarization, long-context retrieval, and free response question-answering?} 
This work answers this question positively by introducing a novel strategy, \ours, that \textit{dynamically} determines token eviction pre-attention via locality-sensitive hashing (LSH)~\citep{goemans1995improved,charikar2002similarity}. 
\ours evicts a past token from the cache whose key embedding is highly cosine dissimilar to the current query token embedding. The intuition behind this strategy is that high cosine dissimilarity indicates a low dot-product attention score. To efficiently scan for cosine (dis)similar tokens without performing attention, \ours leverages the SimHash \citep{charikar2002similarity,goemans1995improved} to instead compare Hamming distances between $c$-length binary hashes of cached key embeddings and the current query embedding. We depict a high-level visualization of this strategy in Figure \ref{visualization-lshe}.

\ours requires minimal overhead: for a total sequence length of $\ell$ tokens with embedding dimension $d$, \ours maintains a constant-size, low-cost binary array in GPU memory of size $c\times k$ bytes, where $c\ll d$ is the hash dimension and $k\ll \ell$. Cached tokens with key embeddings that register low Hamming similarity measurements to decoded query embeddings are gradually replaced.

Our contributions are as follows:
\begin{itemize}[leftmargin=*,itemsep=0pt]
\item \textbf{Novel Attention-Free Token Eviction} We introduce a novel \textit{attention-free} token eviction strategy, \ours, that leverages locality-sensitive hashing (LSH) to quickly locate which token in the cache is the least relevant to the current query. This ranking procedure consists entirely of cheap Hamming distance calculations. The associated binary array for computing these similarities requires minimal memory overhead. For a Llama 3 model, \ours can compress the KV cache by 30\%-70\% with minimal performance drop

\item \textbf{State-of-the-Art Performance} \ours demonstrates high performance on reasoning tasks (GSM8K \cite{cobbe2021training}, MedQA \cite{cobbe2021training}), multiple-choice (GSM8K MC, MedQA MC), long-context retrieval (Needle-in-a-Haystack, Common Word \citep{hsieh2024ruler}), and long-text summarization (MultiNews, GovReport \cite{bai2023longbench}). To the best of our knowledge, \ours achieves state-of-the-art performance for attention-free eviction, outperforming the similar attention-free $L_2$ method. Additionally, \ours outperforms attention-accumulation-based methods on long text summarization tasks and achieves 1.5x speedup in the prefilling stage and comparable speed in the decoding stage withoug low-level optimiations. 

\item \textbf{Open-Source Implementation} Upon public release of our manuscript, we will release an open-source implementation of \ours through a fork of the popular cold-compress library (\url{https://github.com/AnswerDotAI/cold-compress}).
\end{itemize}

\section{Preliminaries}
\label{prelim}
We aim to capture tokens whose query embeddings will form a large sum of dot products (i.e., attention scores) with other key embeddings, but without explicitly calculating attention. We will leverage locality-sensitive hashing (LSH) to quickly determine cosine similarities since the angle is equivalent to the dot product (for unit vectors). In this section, we review technical concepts crucial to attention and locality-sensitive hashing. We assume some base level of similarity with transformers, but we refer the reader to precise formalism \citep{phuong2022formal}. 

\paragraph{Scaled Dot-Product Attention} Consider a sequence of $n$ tokens with $e$-dimensional real-valued representations $x_1, x_2, \dots, x_n$. Let $Q=[q_1 \hspace{.2em} q_2 \hspace{.2em} \cdots \hspace{.2em} q_n]\in\mathbb{R}^{n\times d}$, $K=[k_1 \hspace{.1em} k_2 \hspace{.1em} \cdots \hspace{.2em} k_n]\in\mathbb{R}^{d\times n}$ where $q_i = W_qx_i$, $k_i = W_kx_i$ and $W, K\in \mathbb{R}^{d\times e}$. The query and key projectors $W_q$ and $W_k$ are pre-trained weight matrices. We also define a value matrix $V=[v_1 \hspace{.1em} v_2 \hspace{.1em} v_2 \cdots \hspace{.1em} v_n]\in\mathbb{R}^{d_{out}\times n}$ with $v_i=W_vx_i$ with trainable $V\in\mathbb{R}^{d_{out}\times d}$, the scaled dot-product attention mechanism is given as
\begin{equation}
    \label{attention}
    \textrm{Attention}(Q,K,V)=V\cdot\textrm{softmax}\Bigl(\frac{Q^{\top}K}{\sqrt{d}} \Bigr).
\end{equation}
Typically, attention layers contain multiple heads $\{h_i\}_{i=1}^J$ each with distinct query, key, and value projectors $\{W_q^{(h_i)},W_k^{(h_i)}, W_v^{(h_i)}\}_{i=1}^J$. In a multi-head setup, attention is computed in parallel across all heads, and the outputs are concatenated together and then passed through a linear layer for processing by the next transformer block. 

As $Q, K, V$ are updated with each new incoming token, to avoid significant re-computation, the current state of $Q^{\top}K$, $Q$, and $K$ are maintained in the KV cache. Our goal is to bypass attention computation and caching for select tokens, i.e., sparsify the attention matrix $Q^{\top}K$, $K$, and $V$. 

\paragraph{Locality-Sensitive Hashing} We will now describe a family of locality-sensitive hashing (LSH) functions able to efficiently approximate nearest neighbors (per cosine similarity) of key/query vectors in high-dimensional $\mathbb{R}^d$ through comparison in a reduced $c$-dimensional space (per Hamming distance) with $c\ll d$. Here, "locality-sensitive" means points that are close together according to a distance function $\textrm{dist}_d(\cdot, \cdot)$ in the ambient space remain close per another distance function $\textrm{dist}_c(\cdot, \cdot)$ in the lower-dimensional space with high-probability. For a rigorous treatment of LSH functions, see \citep{andoni2018approximate,charikar2002similarity}. 

Formally for our setup, $dist_d(x,y)\triangleq \cos \theta_{x,y} =\frac{x^{\top}y}{||x||\hspace{0.2em}||y||}$ and $dist_c(p,q) \triangleq \textrm{d}_H(p,q)$ which denotes the Hamming distance. We will project each vector from $\mathbb{R}^d$ into $\mathbb{Z}_2^c$, the space of $c$-bit binary strings (which is often referred to as a \textit{binary hash code}). To acquire a $c$-bit long hash code from an input vector $x \in \R^{d}$, we define a random projection matrix $R \in \R^{c \times d}$ whose entries are independently sampled from the standard normal distribution $\mathcal{N}(0, 1)$. We then define

\begin{equation}
\label{lsh}
    h(x) = \textrm{sgn}(Rx),
\end{equation}
where $\textrm{sgn}(\cdot)$ (as an abuse of conventional notation) is the element-wise Heaviside step function: 
$$ {\displaystyle \textrm{sgn}(x):={\begin{cases}1,&x\geq 0\\0,&x<0\end{cases}}}. $$

For two unit vectors $x,y\in\mathbb{R}^d$ we have that,
\begin{equation}
\label{simhash-expectation}
   \frac{1}{c}\cdot \mathbb{E}[\textrm{d}_H\bigl(h(x),h(y)\bigr)]= \frac{\theta_{x,y}}{\pi},
\end{equation}
where $\theta_{x,y}=\arccos(\cos(\theta_{x,y}))$. We do not prove equation \ref{simhash-expectation} in this work; see Theorem \S 3.1 in \citep[Theorem 3.1]{goemans1995improved}. In particular, if $x$ and $y$ are close in angle, the Hamming distance between $h(x)$ and $h(x)$ is low in expectation. Increasing the hash dimension $c$ reduces variance.

The geometric intuition behind this LSH scheme is the following: each row $R_{:, i}$ of $R$ defines a random hyperplane in $\mathbb{R}^d$. The Heaviside function $\textrm{sgn}(\cdot)$ indicates whether $x$ is positively or negatively oriented with respect to the hyperplane $R_{:, i}$. Thus, the $c$ hyperplanes divide the $d$ dimensional space into multiple partitions, and the resulting $c$-dimensional hash code is an index into one of the partitions in which $x$ is located. Therefore, vectors with the same or similar hash codes lie in the same or close-by partitions and, therefore, are likely similar in angle. 
cwecwasdf

\subsection{Related Works}

\paragraph{KV Cache Compression} 
Many popular compression strategies adopt an \textit{eviction} approach, which removes embeddings from the KV cache. H$_2$O \citep{zhang2024h2o} and Scissorhands \citep{liu2024scissorhands} calculate token importance by their accumulated attention scores and keep the ``heavy hitters" in the cache. FastGen \citep{ge2023model} performs a profiling pass before the generation stage that assigns to each head, according to the head's attention patterns, a pruning policy which only retains categories of tokens (punctuation, special, etc.) favored by the head. These eviction strategies depend on the computation of attention scores for their policy. An attention-free $L_2$ dropout method \citep{devoto2024simple}, which we compare ourselves to in this work, uses the observation that high-attention tokens tend to have low $L_2$ key norms to approximately keep important tokens in cache. Other methods seek to merge KV caches across heads, such as grouped query attention (GQA) \citep{ainslie2023gqa, dubey2024llama}. KVMerger \citep{wang2024model} and MiniCache \citep{liu2024minicache}, which searches for similarity between tokens in consecutive attention layers and subsequently merges KV cache entries across these layers. While these consolidation approaches prevent memory complexity associated with KV caches from scaling with depth or multi-head attention, the size of any singular cache still tends to scale with sequence length. 

\paragraph{LSH Based Attention}
Similar to our work, Reformer \citep{kitaev2020reformer} employs LSH to find similar tokens, but as a way to replace the softmax attention as opposed to token eviction. It creates hash buckets of tokens that form local attention groups and only attends to tokens in the same and neighboring buckets. However, this makes Reformer vulnerable to missing important tokens due to hash collision or boundary issues, and therefore, it must use multiple hash tables to mitigate this issue. In a similar vein, KDEFormer \citep{zandieh2023kdeformer}, HyperAttention \citep{han2023hyperattention}, and \cite{zandieh2024qjl}, use LSH to stably approximate and compressing the attention module thus accelerating the computation, but without token eviction. SubGen \citep{zandieh2024subgen} uses LSH to cluster key embeddings and samples representatives from each cluster to reduce the size of the KV Cache and consequently speed up attention, though it must initially view all queries and keys to perform this clustering which could result in VRAM blowup, which our method avoids.

\paragraph{Memory Efficient Transformers}
Multi-Query Attention \citep{shazeer2019fast} and Grouped Query Attention \citep{ainslie2023gqa} reduce the number of key-value matrices by sharing them across multiple query heads to save KV cache memory usage. However, they require retraining or up-training the LLM. Cache quantization methods \citep{hooper2024kvquant,sheng2023flexgen} reduce the KV cache size by compressing the hidden dimension instead of along the sequence dimension but can result in information loss. Linear Transformer \citep{katharopoulos2020transformers} reduces memory usage by replacing softmax attention with linear kernels and, therefore, achieves a constant memory requirement.

\section{\ours: A Locality-Sensitive Eviction Strategy}
\label{algo}
We now formalize our eviction method reflected in Algorithm \ref{alg:cap}. We assume that the KV cache has a limited and fixed budget and conceptually divide the KV cache management during LLM inference into two stages: the initial Prompt Encoding Stage and then a Decoding Stage (i.e., generation).

Let $C$ be a constant and fixed cache budget, $\mathcal{K}$ be the key cache, and $\mathcal{V}$ be the V cache in a K-V attention head. We define our eviction policy as a function 
\begin{equation}
\label{policy}
    \mathcal{K}_t, \mathcal{V}_t, \mathcal{H}_t \leftarrow P(q, \mathcal{K}_{t-1}, \mathcal{V}_{t-1}, \mathcal{H}_{t-1})
\end{equation}
where $\mathcal{H}_t \in \{0,1\}^{b \times C}$ is a hash table that contains hash codes of keys in $\mathcal{K}$. We then define a function $F_{score}$ to assign a score for each key inside the K cache. $F_{score}$ outputs an array which contains the negative of hamming distances $d_H$ between the hash code of a query vector $q$ and columns of $\mathcal{H}$, which are hash codes of all non-evicted keys. 
\begin{equation}
\label{score_fn}
    F_{score}(q,\mathcal{K}) = -d_H(h(q),\mathcal{H})
\end{equation}
The eviction index $e_t$ at any step $t$ is selected as the index with the lowest score:
\begin{equation}
\label{evict_index}
    e_t \leftarrow \mathop{\arg \min}  F_{score}(q_{t-1}, \mathcal{H}_{t-1})
\end{equation}
which points to the key that is most distant from the query vector at time step $t$. Entries at index $e_t$ from the $\mathcal{K}$ and $\mathcal{V}$ are evicted and $\mathcal{H}$ is updated (step 3-6 of Algorithm \ref{alg:cap}). 

\begin{algorithm}
\caption{\ours(timestep $t$)}\label{alg:cap}
\begin{algorithmic}[1]
\Require query $q$, key $k$, value $v$, key cache $\mathcal{K}$, value cache $\mathcal{V}$, hash table $\mathcal{H}$ 
\State $e_t \leftarrow \mathop{\arg \min} F_{score}(q_t, \mathcal{H}_{t-1})$ \Comment{Determine eviction index $e_t$}
\State \textbf{del} $\mathcal{K}_{t-1}^{e_t}$, $\mathcal{V}_{t-1}^{e_t}$,  $\mathcal{H}_{t-1}^{e_t}$ \Comment{Remove entries at index $e_t$ from KV cache and hash table}
\State $\mathcal{K}_t \leftarrow \mathcal{K}_{t-1} \cup k_t$ \Comment{Update key cache}
\State $\mathcal{V}_t \leftarrow \mathcal{V}_{t-1} \cup v_t$  \Comment{Update value cache}
\State $\mathcal{H}_t \leftarrow \mathcal{H}_{t-1} \cup h(k_t)$ \Comment{Add hash of $k_t$ to the hash table}
\State $A \leftarrow \text{Attention}(q, \mathcal{K}_T, \mathcal{V}_T)$ \Comment{Calculate attention}
\end{algorithmic}
\end{algorithm}

\paragraph{Prompt Encoding Stage}
During the prompt encoding stage, the model processes the prompt, $x_{prompt} = [x_1, ... , x_N] \in \R^{N \times d}$. The KV cache and the hash table are first filled to full by the first $C$ tokens. $\mathcal{K}_0 = \{k_1, ..., k_C\}, \mathcal{V}_0 = \{v_1, ..., v_C\}, \mathcal{H}_0 = h(\mathcal{K}_0) = \bigcup_{i\in [1, C]} h(k_i)$. We then set $t\leftarrow C+1$, and begin Algorithm \ref{alg:cap}.

\paragraph{Decoding Stage}
Let $x_{decoding} = [z_{1}, ... z_T] \in \R^{T \times d}$ be the generated tokens during auto-regressive decoding. In the decoding stage, we continue Algorithm \ref{alg:cap} by setting $t <- N+1$. The generation completes at time step $N+T$.

\paragraph{Complexity} Our strategy assumes a fixed memory budget, and therefore, uses constant memory. The computation overhead per time step is also constant, because $F_score$ is calculated for a constant $C$ number of key vectors in the cache. The extra memory overhead that \ours  introduces to each attention head is the hash table $\mathcal{H}$, which only uses $C * b$ bits of space and is independent of the sequence length. The hash table is stored on GPU memory and does not introduce any latency bottlenecks associated with CPU-to-GPU streaming \citep{strati2024d}.

\section{Experiments}
\paragraph{Tasks} 
\label{exps}
We evaluated \ours across various tasks to demonstrate its effectiveness in reducing the memory cost of the KV cache while preserving the language quality of the generated text. Our experiments are split into four main categories: free response question answering, multiple choice, long-context retrieval and long-context summarization. Our long context retrieval tasks include the multi-key needle-in-a-haystack task and the common words task from \citep{hsieh2024ruler}. Question answering tasks include GSM8K \citep{cobbe2021training} and MedQA \citep{jin2021disease}. Summarizaiton tasks include GovReport and MultiNews from \cite{bai2023longbench}. 

\paragraph{Metrics} 
The question-answering tasks were evaluated using BERTScore (which includes precision, recall, and F1 scores), ROUGE (ROUGE-1, ROUGE-2 and ROUGE-L and ROUGE-Lsum), and GPT4-Judge. GPT-4 was prompted to look at both the model prediction and the ground truth answer, then provide a score from 1 - 5 on the coherence, faithfulness, and helpfulness of the answer in addition to similarity between the prediction and ground truth (we named this metric GPT4-Rouge). In this section, we report the average of these four scores. For details on individual scores, please see Appendix \ref{detail-results}. For the system prompts given to GPT-4, refer to Appendix \ref{appendix:prompt}. For multiple-choice tasks, we use accuracy as our metric. The metric used to evaluate long context retrieval tasks is the string matching score from \cite{hsieh2024ruler}, whose definition is in Appendix \ref{appendix:string-match-score}. For summarization tasks, we use Rouge as the metric as per direction from \cite{bai2023longbench}.

\paragraph{Configuration and Setup} We conducted most experiments using Meta's Llama3 8B-Instruct model \citep{dubey2024llama} with the exception of long text summarization tasks which were tested using the Llama3.1 8B-Instruct model. Our method is agnostic to grouped-query attention, so we used the default group size of 4. The maximum sequence length was set to the sum of the maximum prompt length and the maximum number of allowed generated tokens needed for each task. We conducted experiments using cache budgets of 10\%, 30\%, 50\%, 70\%, and 90\% of the full KV cache. Based on insights from \citep{xiao2023efficient, child2019generating, beltagy2020longformer}, we also keep the most recent 10 tokens and the first 4 tokens of the prompt always in the KV cache. The summarization tasks were performed on Nvidia H100 80GB graphics cards due to their long contexts. All other experiments were conducted on the Google Cloud Platform G2 instances with Nvidia L4 24GB graphics cards. 

\paragraph{Baseline Methods} We chose the $L_2$ norm-based eviction method \citep{devoto2024simple} as our main baseline for comparison because it is also an attention-free KV cache eviction method. We also included two attention-accumulation-based methods: H$_2$O \cite{zhang2024h2o} and Scissorhands \cite{liu2024scissorhands}, as well as a hybrid method: Fastgen \cite{ge2023model}.

\subsection{Free Response Question Answering}
We tested our strategy against tasks that require generating accurate answers using multi-step reasoning. Specifically, we used the GSM8K and MedQA datasets to assess language quality for each strategy, given a constrained KV cache budget. Both tasks are used to test the potential side effects of compression on the LLM's reasoning ability. 

\begin{figure}[htbp]
    \begin{subfigure}[t]{0.33\textwidth}
         \centering
        \includegraphics[width=\linewidth]{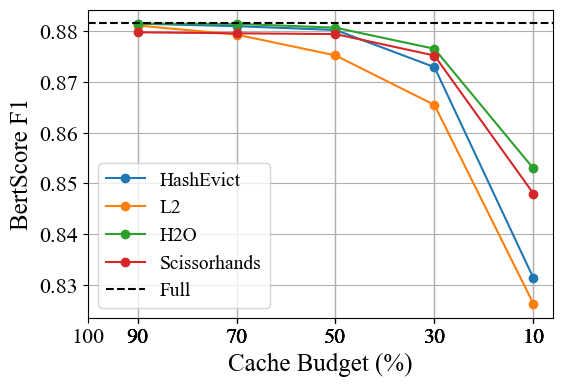}
        \caption{BERTScore F1}
        \label{fig:gsm-f1}
    \end{subfigure}
    \begin{subfigure}[t]{0.33\textwidth}
        \centering
        \includegraphics[width=\linewidth]{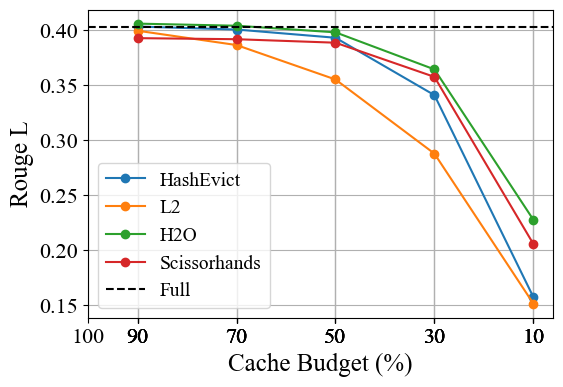}
        \caption{Rouge L}
        \label{fig:gsm-rougel} 
    \end{subfigure}
    \begin{subfigure}[t]{0.33\textwidth}
        \centering
        \includegraphics[width=\linewidth]{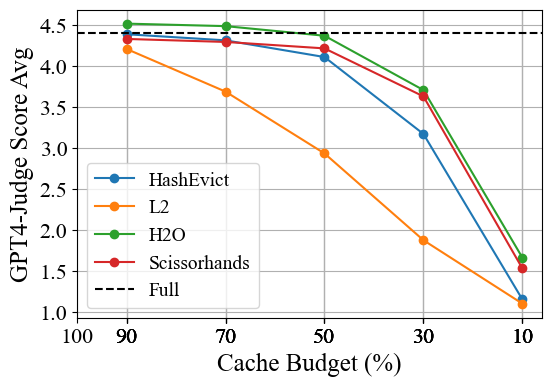}
        \caption{GPT4-Judge}
        \label{fig:gsm-gpt4}
    \end{subfigure}
\caption{\textbf{GSM8K Question Answering  Performance.} We measure BERTScore F1, Rouge-L, and GPT4-Judge for different cache budgets on a grade school math task. \ours outperforms $L_2$ for all three metrics for every budget, with sharp differences for the 50\% and 30\% compression. \ours performs similarly to H$_2$O and Scissorhands except at 10\% cache budget.}
\label{fig:gsm-free-form}
\end{figure}

\paragraph{GSM8K}
GSM8K consists of grade-school-level math problems that typically require multiple reasoning steps. As shown in Figure \ref{fig:gsm-free-form}, \ours strategy consistently outperforms the $L_2$ norm-based method across various cache sizes. Notably, even when the KV cache budget is set to 50\% of the full capacity, the \ours eviction strategy maintains a high answer quality, with minimal degradation in BERTScore F1, ROUGE-L, and GPT4-Judge scores. Additionally, \ours performs on par with H2O and Scissorhands without accumulating attention scores. 

\begin{figure}[htbp]
    \begin{subfigure}[t]{0.33\textwidth}
         \centering
        \includegraphics[width=\linewidth]{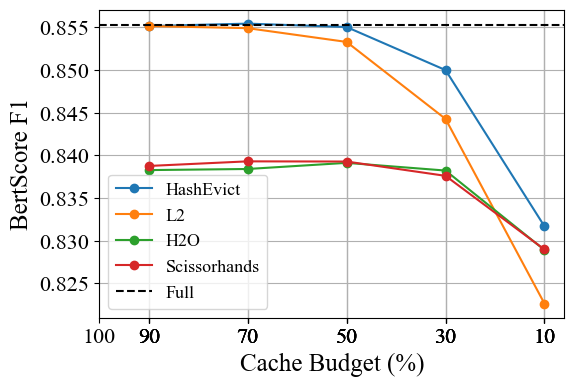}
        \caption{BERTScore F1}
        \label{fig:medqa-f1}
    \end{subfigure}
    \begin{subfigure}[t]{0.33\textwidth}
        \centering
        \includegraphics[width=\linewidth]{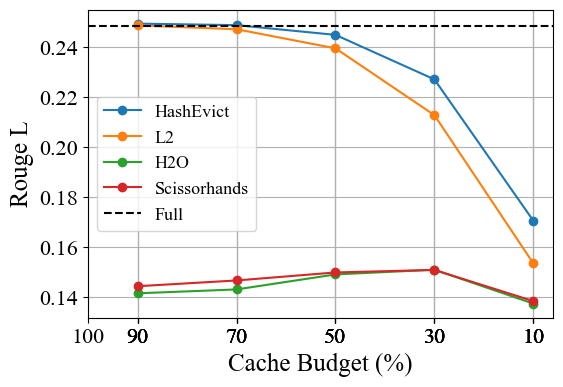}
        \caption{Rouge L}
        \label{fig:medqa-rougel} 
    \end{subfigure}
    \begin{subfigure}[t]{0.33\textwidth}
        \centering
        \includegraphics[width=\linewidth]{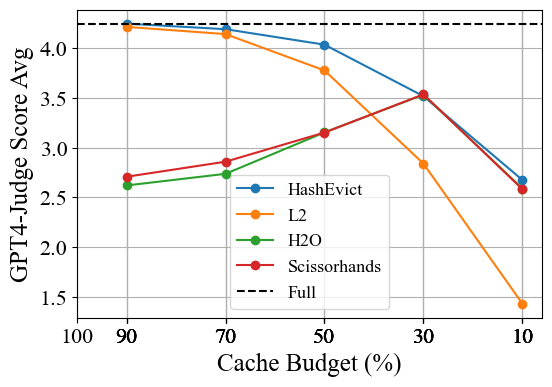}
        \caption{GPT4-Judge}
        \label{fig:medqa-gpt4}
    \end{subfigure}
\caption{\textbf{MedQA Question Answering Performance.} We measure BertScore F1, Rouge-L, and GPT4-Judge for different cache budgets on a medical exam task. LSH outperforms $L_2$ for all three metrics for every budget, with a significantly higher performance for the 30\% and 10\% budgets. }
\label{fig:medqa-free-form}
\end{figure}

\paragraph{MedQA}
MedQA is a free response multiple choice question answering dataset collected from professional medical board exams. We randomly sampled 100 questions from this dataset. Each question has 5 choices and only one correct answer, along with ground truth explanations and reasoning steps. Figure \ref{fig:medqa-free-form} illustrates that \ours performs better than all baseline methods for all cache budgets tested. For both datasets, \ours produced more coherent and helpful answers across all cache budgets than the baselines per Table \ref{MedQA-qa}.

For detailed experiment results of both question anwering tasks, and for comparison with Fastgen at various attention recovery ratios, please refer to Appendix \ref{detail-results}.

\subsection{Multiple Choice Question Answering}
We evaluated our method on multiple-choice versions of GSM8K and MedQA. Multiple choice is a more difficult test of a model's reasoning capability under the constraint of cache compression, as it takes away the ability to use intermediate results in the generated text. The model has to keep useful tokens during prompt compression in order to pick the correct answer choice. 

\begin{figure}[ht]
    \centering
    \begin{subfigure}[t]{0.48\textwidth} 
        \centering
        \includegraphics[width=\linewidth]{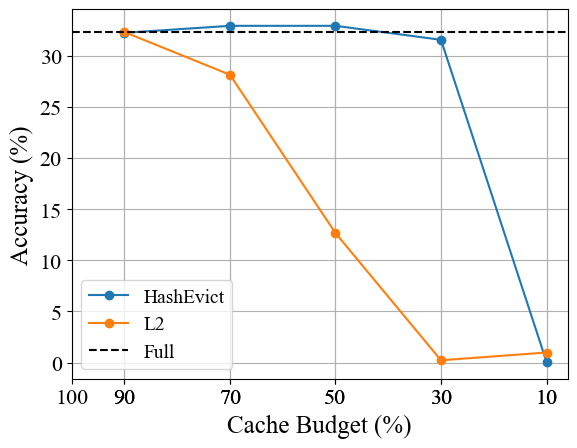} 
        \caption{Accuracy on the GSM8K Multiple Choice}
        \label{fig:gsm-mc}
    \end{subfigure}\hspace{0.02\textwidth}
    \begin{subfigure}[t]{0.48\textwidth} 
        \centering
        \includegraphics[width=\linewidth]{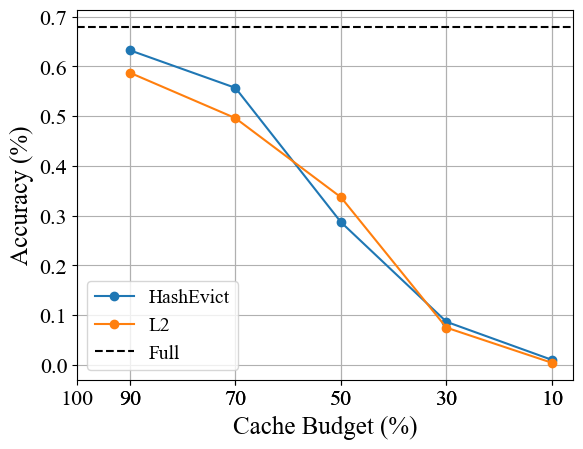} 
        \caption{Accuracy on the MedQA Multiple Choice}
        \label{fig:medqa-mc}
    \end{subfigure}
    \caption{\textbf{Multiple Choice Tasks Performance.} On GSM8K, \ours outperforms the baseline full cache on GSM8K at 70\% and 50\% cache budgets and significantly outperforms $L_2$ at 70\%, 50\%, and 30\%. \ours performs on par with $L_2$ overall on MedQA with higher performance at 90\% (near uncompressed performance) and 70\% budget and slightly lower performance at 50\% budget.}
    \label{fig:mc}
\end{figure}

\paragraph{GSM8K Multiple Choice}
For the GSM8K multiple choice experiments, \ours significantly outperforms $L_2$ for cache budgets of 30\% and 50\%. As shown in Figure \ref{fig:gsm-mc}, the $L_2$ method's accuracy drops significantly at smaller cache sizes, while the performance of \ours does not significantly drop until the cache budget is set at 10\%.

\paragraph{MedQA Multiple Choice}
Per Figure \ref{fig:medqa-mc}, the MedQA multiple choice experiment, \ours offers better performance than $L_2$ eviction for all tested cache budgets except for 50\%. Performance between both methods is highly similar at lower budgets.

\subsection{Long-Context Retrieval}
To evaluate \ours's ability to retain and retrieve important pieces of information from long contexts, we used the Needle-in-a-Haystack and Common Words tasks from \cite{hsieh2024ruler} with 4K context length. These tests benchmark the ability of a compression strategy to retain important tokens inside the KV cache within a large, complext stream of context.

\begin{figure}[htbp]
    \centering
    \begin{subfigure}[t]{0.48\textwidth}
        \centering
        \includegraphics[width=\linewidth]{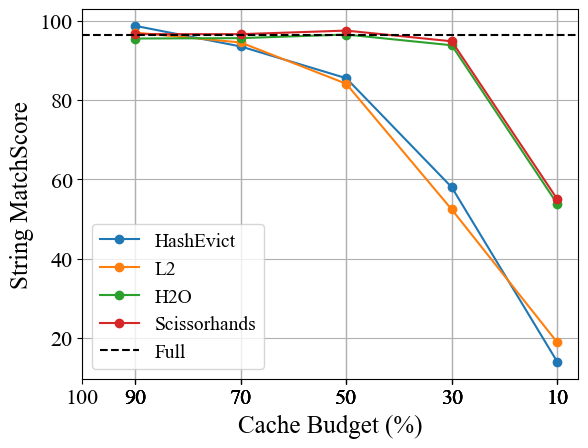} 
        \caption{String Match Score on Common Words}
        \label{fig:cwe}
    \end{subfigure} \hspace{0.02\textwidth}
    \begin{subfigure}[t]{0.48\textwidth} 
        \centering
        \includegraphics[width=\linewidth]{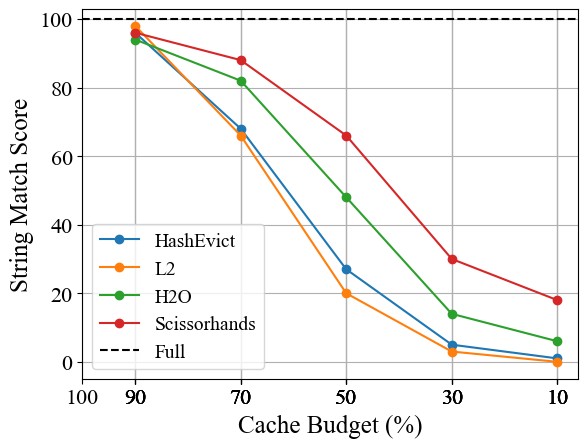}
        \caption{String Match Score on Needle-in-a-Haystack}
        \label{fig:niah}
    \end{subfigure}
\caption{\textbf{Long-Context Tasks.} We measure string-matching scores for two long-context retrieval tasks. \ours performs on par with $L_2$ on the Common Words task with slightly higher performance at a 30\% cache budget and slightly lower performance at a 10\% budget. For the Needle-in-a-Haystack task, \ours performs on par with $L_2$ with slightly higher performance at a 50\% cache budget.}
\label{fig:long-context}
\end{figure}

\paragraph{Needle-in-a-Haystack}
In the Needle-in-a-Haystack task, the model must extract specific information buried within a large body of text. As illustrated in Figure \ref{fig:niah}, \ours slightly outperforms $L_2$ at every cache budget except for 90\%, and both methods see a sharp drop in the ability to recall the ``needle" (a small, targeted piece of context) after the cache budget drops to 50\% and lower. \ours outperforms $L_2$ for these smaller cache sizes.

\paragraph{Common Words}
In the Common Words task, the model must identify the most frequent words from a long list. Figure \ref{fig:cwe} demonstrates that \ours performs on par with $L_2$ eviction in general and slightly better at 30\%, 50\%, and 90\% cache budget. Both methods outperform the full cache model at 90\% cache size, indicating that some cache compression can actually increase performance. Neither method experienced a significant drop in performance until the cache budget was reduced to 30\%.

\subsection{Long Text Summarization}
To evaluate \ours's ability to handle exceptionally long context lengths, we incorporated the MultiNews and GovReport summarizations tasks from LongBench \cite{bai2023longbench}. We tested both tasks using the Llama3.1-8B-Instruct model and used context size of 16K tokens.

\begin{figure}[htbp]
    \begin{subfigure}[t]{0.48\textwidth}
         \centering
        \includegraphics[width=\linewidth]{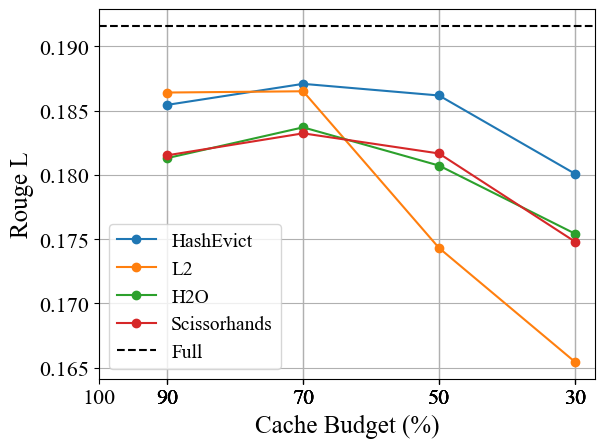}
        \caption{Rouge-L on MultiNews}
        \label{fig:longbench-multinews}
    \end{subfigure}
    \begin{subfigure}[t]{0.48\textwidth}
        \centering
        \includegraphics[width=\linewidth]{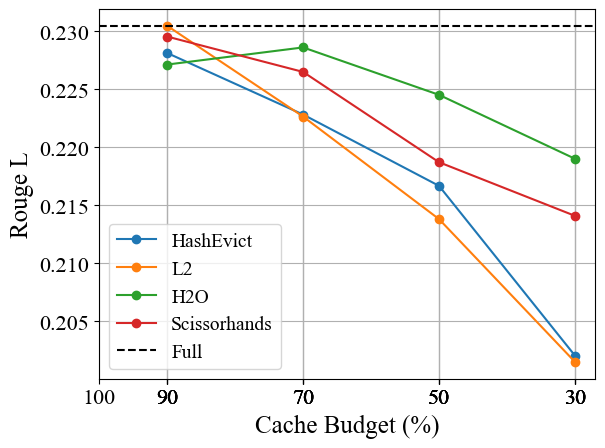}
        \caption{Rouge-L on GovReport}
        \label{fig:longbench-govreport} 
    \end{subfigure}
\caption{\textbf{LongBench Summarization Tasks} We measure Rouge-L for two long text summarization tasks. \ours outperforms all baseline methods on MultiNews at 30 - 70\% cache budget. \ours performs better than $L_2$ on GovReport at 50\% cache budget similarly at 30\% and 70\%. }
\label{fig:longbench}
\end{figure}

\paragraph{MultiNews}
The MultiNews dataset contains clusters of 2-10 news articles discussing the same event or topic. The model is asked to provide a one-page summary of the articles. \ours outperforms all baselines in the MultiNews summarization task at 30-70\% cache budget. At 90\% cache budget, \ours still outperforms H$_2$O and Scissorhands while being slighly lower thant $L_2$. 

\paragraph{GovReport} 
The GovReport dataset contains reports spanning a wide variety of national policy issues from the U.S. Government. The model is asked to produce a one-page summary of the reports. \ours performs on par with and sometimes slightly better than $L_2$ at 30-70\% cache budget, while not as well as H$_2$O or Scissorhands.

\subsection{Throughput}
To evaluate the speed of \ours and baseline methods, we measured the decoding and prefilling speed during the MultiNews evaluation. Because the length of answers generated by each eviction strategy generates can be different, we report decoding and prefilling speed in tokens per second instead of elapsed time. 

\begin{table}[htbp]
\centering
\caption{\textbf{Throughput on LongBench MultiNews Summarization Task} \ours method is as fast as $L_2$ and faster than other baselines at both prefilling and decoding, even without low-level optimizations (i.e., expressing our hash tables in true binary bits). At the prefill stage, \ours is 1.5x as fast as H$_2$O and Scissorhands and 17x as fast compared to FastGen. }

\begin{tabular}{@{}ccccc@{}}
\toprule
\begin{tabular}[c]{@{}c@{}}Cache Budget (\%)\\ / Fastgen Attn \\ Recovery Frac (\%)\end{tabular} & Strategy & Rouge L & \begin{tabular}[c]{@{}c@{}}Decode\\ Toks Per Sec\end{tabular} & \begin{tabular}[c]{@{}c@{}}Prefill\\ Tokes Per Sec\end{tabular} \\ \midrule
\multirow{4}{*}{30} & \multicolumn{1}{c|}{\ours} & 0.180 & 22.880 & 20293.524 \\
 & \multicolumn{1}{c|}{$L_2$} & 0.165 & 23.981 & 20628.160 \\
 & \multicolumn{1}{c|}{H$_2$O} & 0.175 & 21.555 & 13025.776 \\
 & \multicolumn{1}{c|}{Scissorhands} & 0.175 & 21.448 & 13004.254 \\ \midrule
\multirow{4}{*}{50} & \multicolumn{1}{c|}{\ours} & 0.186 & 22.846 & 20459.961 \\
 & \multicolumn{1}{c|}{$L_2$} & 0.174 & 16.013 & 15851.952 \\
 & \multicolumn{1}{c|}{H$_2$O} & 0.181 & 21.973 & 13969.985 \\
 & \multicolumn{1}{c|}{Scissorhands} & 0.182 & 20.978 & 13549.967 \\ \midrule
\multirow{4}{*}{70} & \multicolumn{1}{c|}{\ours} & 0.187 & 22.914 & 21002.334 \\
 & \multicolumn{1}{c|}{$L_2$} & 0.187 & 24.305 & 21303.763 \\
 & \multicolumn{1}{c|}{H$_2$O} & 0.184 & 21.793 & 14050.521 \\
 & \multicolumn{1}{c|}{Scissorhands} & 0.183 & 21.705 & 13954.693 \\ \midrule
\multirow{4}{*}{90} & \multicolumn{1}{c|}{\ours} & 0.185 & 22.873 & 21229.230 \\
 & \multicolumn{1}{c|}{$L_2$} & 0.186 & 24.010 & 21305.693 \\
 & \multicolumn{1}{c|}{H$_2$O} & 0.181 & 21.665 & 14007.697 \\
 & \multicolumn{1}{c|}{Scissorhands} & 0.182 & 21.411 & 14025.440 \\ \midrule
100 & \multicolumn{1}{c|}{Full} & 0.192 & 16.071 & 16573.492 \\ \midrule
70 & \multicolumn{1}{c|}{\multirow{4}{*}{Fastgen}} & 0.129 & 12.752 & 1171.069 \\
75 & \multicolumn{1}{c|}{} & 0.174 & 12.291 & 1157.987 \\
80 & \multicolumn{1}{c|}{} & 0.184 & 11.850 & 1142.679 \\
85 & \multicolumn{1}{c|}{} & 0.183 & 11.658 & 1164.689 \\ \bottomrule
\end{tabular}

\label{tab:throughput}
\end{table}

\subsection{Memory Usage}
Table \ref{table:cache-mem-usage} compares the memory usage of the KV cache and relevant data structures of $L_2$ and \ours  on the GSM8K and MedQA question answering experiments. \ours  maintains $\mathcal{H}$, a binary hash matrix of the attention keys in memory and, therefore, has slightly higher memory usage than $L_2$ eviction. Our implementation uses 8 bits for binary values instead of 1 bit. Using 1-bit binary numbers would reduce the memory overhead of \ours by a factor of 8 and narrow the difference in memory usage between \ours and $L_2$.

\begin{table}[htbp]
\caption{\textbf{GSM8K and MedQA Question Answering KV Cache Memory Usage.} \ours maintains a binary hash matrix of attention keys in memory and, therefore, has slightly higher memory usage than $L_2$. Our implementation uses 8-bits for binary values instead of 1-bit. Using 1-bit binary numbers will reduce the memory overhead of \ours by a factor of 8 and decrease the difference in memory usage between \ours and $L_2$.}
\centering
\begin{tabular}{@{}cc|cc|cc@{}}
\toprule
 &  & \multicolumn{2}{c|}{GSM8K} & \multicolumn{2}{c}{MedQA} \\ \midrule
\begin{tabular}[c]{@{}c@{}}Cache\\ Budget\\ (\%)\end{tabular} & Strategy & \begin{tabular}[c]{@{}c@{}}Compression\\ Ratio\end{tabular} & \begin{tabular}[c]{@{}c@{}}Cache\\ Memory\\ (GB)\end{tabular} & \begin{tabular}[c]{@{}c@{}}Compression\\ Ratio\end{tabular} & \begin{tabular}[c]{@{}c@{}}Cache\\ Memory\\ (GB)\end{tabular} \\ \midrule
\multirow{2}{*}{10} & $L_2$ & 0.8355 & 0.7603 & 0.9289 & 2.5342 \\
 & \ours & 0.8380 & 0.8120 & 0.8812 & 2.6338 \\ \midrule
\multirow{2}{*}{30} & $L_2$ & 0.6234 & 1.7740 & 0.6957 & 7.3492 \\
 & \ours & 0.6018 & 1.8531 & 0.6360 & 7.5786 \\ \midrule
\multirow{2}{*}{50} & $L_2$ & 0.3968 & 2.7876 & 0.4175 & 12.1641 \\
 & \ours & 0.3716 & 2.8941 & 0.3901 & 12.5235 \\ \midrule
\multirow{2}{*}{70} & $L_2$ & 0.1967 & 3.8013 & 0.1803 & 17.2325 \\
 & \ours & 0.1857 & 3.9351 & 0.1740 & 17.7285 \\ \midrule
\multirow{2}{*}{90} & $L_2$ & 0.0859 & 4.8150 & 0.0498 & 22.0474 \\
 & \ours & 0.0823 & 4.9761 & 0.0483 & 22.6734 \\ \midrule
100 & Full & 0.0000 & 12.6934 & 0.0000 & 51.1181 \\ \bottomrule
\end{tabular}
\label{table:cache-mem-usage}
\end{table}

\subsection{Ablation on Hash Dimension}

To determine the effect of the hash dimension, we conducted an ablation study using the GSM8K free response dataset. Fixing the cache budget to 50\%, we tested hash dimensions of 4, 8, 16, 32 and 64 bits.  The choice of hash dimension does not significantly impact performance. In fact, 8 bits performed the best, but not noticeably better than higher dimensions. This demonstrates that \ours does not require a high hashing dimension and can be executed with minimal storage overhead. When using 8 bits, the storage overhead is 1 byte $\times$ cache size. For example, in a Llama3 70B-Instruct deployment with 80 layers, 8 KV-heads, sequence length of 8192, batch size of 8 and 50\% cache budget, hash dimension of 8-bits, we have that 16-bits and 32-bits only use an extra 20MB, 40MB, and 80MB respectively, which are significantly smaller than the KV cache size of 640GB. Detailed results can be found in Table \ref{table:ablation} of Appendix \ref{appendix:lsh-dim-ablation}.

\subsection{Attention Loss Ratio}

\begin{table}[htbp]
\centering
\caption{\textbf{Attention Loss Ratio} measured at 50\% cache budget on the GSM8K question answering dataset. \ours achieves lower attention loss ratio compared to $L_2$ and even the attention based Scissorhands.}
\begin{tabular}{@{}c|c@{}}
\toprule
Strategy & Attention Loss \\ \midrule
\ours & 0.0336 \\
$L_2$ & 0.0340 \\
Scissorhands & 0.0448 \\
H$_2$O & 0.0139 \\ \bottomrule
\end{tabular}
\label{table:attention_loss}
\end{table}

To empirically verify that \ours selects and evicts tokens of low relevance, we measured the average attention loss of the attention heads for \ours and compared with $L_2$, Scissorhands and H$_2$O. Attention loss is defined as the sum of the attention probabilities for the evicted tokens. Or equivalently, 1 - the sum of the attention probabilities for the tokens in the compressed cache. The attention loss ratios were measured at 50\% cache budget using prompts from the GSM8K question answering dataset. As per table \ref{table:attention_loss}, by accumulating attention score of tokens H$_2$O has the lowest attention loss ratio as expected. \ours has slightly lower attention loss compared to $L_2$ and even the attention-based Scissorhands, proving \ours's ability of discarding irrelevant tokens.

\section{Discussion \& Conclusion} \label{sec:conc}

In this paper, we introduce \ours, a novel attention-free eviction strategy for KV cache compression in transformer-based LLMs. By leveraging locality-sensitive hashing (LSH) to approximate cosine similarity, \ours dynamically determines which tokens to evict from the cache without performing costly attention calculations. Our experiments demonstrate that \ours can achieve 30-70\% compression of the KV cache while maintaining strong performance across various tasks, including free-response Q\&A, multiple-choice Q\&A, and long-context retrieval.

The key advantage of \ours lies in its ability to efficiently compress the KV cache pre-attention, enabling significant memory savings and faster inference times. Compared to traditional strategies like $L_2$ norm-based eviction \citep{devoto2024simple}, \ours excels particularly in reasoning and multiple-choice tasks, where maintaining a diverse set of tokens in the cache is crucial for generating accurate and coherent responses.

There are several potential areas for future work. Investigating hybrid approaches that combine LSH-based eviction with attention-based mechanisms such as \citep{zhang2024h2o, ge2023model} could offer a middle ground between computational efficiency and retention of high-importance tokens. Further, reducing the overhead associated with maintaining binary hash codes (e.g., by optimizing bit precision) could further enhance the applicability of \ours to memory-constrained environments.

\section*{Acknowledgments}
Liu and Fermüller are supported by the NSF under award BCS 2318255 and the National MS Society under award RG-2110-38460. Huang is supported by DARPA Transfer from Imprecise and Abstract Models to Autonomous Technologies (TIAMAT) 80321, National Science Foundation NSF-IIS-2147276 FAI, DOD-AFOSR-Air Force Office of Scientific Research under award number FA9550-23-1-0048, Adobe, Capital One and JP Morgan faculty fellowships.

\bibliographystyle{unsrtnat}
\bibliography{references}

\begin{thebibliography}{39}
\providecommand{\natexlab}[1]{#1}
\providecommand{\url}[1]{\texttt{#1}}
\expandafter\ifx\csname urlstyle\endcsname\relax
  \providecommand{\doi}[1]{doi: #1}\else
  \providecommand{\doi}{doi: \begingroup \urlstyle{rm}\Url}\fi

\bibitem[Zhao et~al.(2023)Zhao, Zhou, Li, Tang, Wang, Hou, Min, Zhang, Zhang, Dong, et~al.]{zhao2023survey}
Wayne~Xin Zhao, Kun Zhou, Junyi Li, Tianyi Tang, Xiaolei Wang, Yupeng Hou, Yingqian Min, Beichen Zhang, Junjie Zhang, Zican Dong, et~al.
\newblock A survey of large language models.
\newblock \emph{arXiv preprint arXiv:2303.18223}, 2023.

\bibitem[Wei et~al.(2022)Wei, Tay, Bommasani, Raffel, Zoph, Borgeaud, Yogatama, Bosma, Zhou, Metzler, et~al.]{wei2022emergent}
Jason Wei, Yi~Tay, Rishi Bommasani, Colin Raffel, Barret Zoph, Sebastian Borgeaud, Dani Yogatama, Maarten Bosma, Denny Zhou, Donald Metzler, et~al.
\newblock Emergent abilities of large language models.
\newblock \emph{arXiv preprint arXiv:2206.07682}, 2022.

\bibitem[Bahdanau(2014)]{bahdanau2014neural}
Dzmitry Bahdanau.
\newblock Neural machine translation by jointly learning to align and translate.
\newblock \emph{arXiv preprint arXiv:1409.0473}, 2014.

\bibitem[Luong(2015)]{luong2015effective}
Minh-Thang Luong.
\newblock Effective approaches to attention-based neural machine translation.
\newblock \emph{arXiv preprint arXiv:1508.04025}, 2015.

\bibitem[Vaswani(2017)]{vaswani2017attention}
A~Vaswani.
\newblock Attention is all you need.
\newblock \emph{Advances in Neural Information Processing Systems}, 2017.

\bibitem[Dubey et~al.(2024)Dubey, Jauhri, Pandey, Kadian, Al-Dahle, Letman, Mathur, Schelten, Yang, Fan, et~al.]{dubey2024llama}
Abhimanyu Dubey, Abhinav Jauhri, Abhinav Pandey, Abhishek Kadian, Ahmad Al-Dahle, Aiesha Letman, Akhil Mathur, Alan Schelten, Amy Yang, Angela Fan, et~al.
\newblock The llama 3 herd of models.
\newblock \emph{arXiv preprint arXiv:2407.21783}, 2024.

\bibitem[Fu(2024)]{fu2024challenges}
Yao Fu.
\newblock Challenges in deploying long-context transformers: A theoretical peak performance analysis.
\newblock \emph{arXiv preprint arXiv:2405.08944}, 2024.

\bibitem[Ge et~al.(2023)Ge, Zhang, Liu, Zhang, Han, and Gao]{ge2023model}
Suyu Ge, Yunan Zhang, Liyuan Liu, Minjia Zhang, Jiawei Han, and Jianfeng Gao.
\newblock Model tells you what to discard: Adaptive kv cache compression for llms.
\newblock \emph{arXiv preprint arXiv:2310.01801}, 2023.

\bibitem[Zhang et~al.(2024{\natexlab{a}})Zhang, Sheng, Zhou, Chen, Zheng, Cai, Song, Tian, R{\'e}, Barrett, et~al.]{zhang2024h2o}
Zhenyu Zhang, Ying Sheng, Tianyi Zhou, Tianlong Chen, Lianmin Zheng, Ruisi Cai, Zhao Song, Yuandong Tian, Christopher R{\'e}, Clark Barrett, et~al.
\newblock H2o: Heavy-hitter oracle for efficient generative inference of large language models.
\newblock \emph{Advances in Neural Information Processing Systems}, 36, 2024{\natexlab{a}}.

\bibitem[Zhang et~al.(2024{\natexlab{b}})Zhang, Liu, Chen, Kailkhura, Chen, and Wang]{zhang2024q}
Zhenyu Zhang, Shiwei Liu, Runjin Chen, Bhavya Kailkhura, Beidi Chen, and Atlas Wang.
\newblock Q-hitter: A better token oracle for efficient llm inference via sparse-quantized kv cache.
\newblock \emph{Proceedings of Machine Learning and Systems}, 6:\penalty0 381--394, 2024{\natexlab{b}}.

\bibitem[Devoto et~al.(2024)Devoto, Zhao, Scardapane, and Minervini]{devoto2024simple}
Alessio Devoto, Yu~Zhao, Simone Scardapane, and Pasquale Minervini.
\newblock A simple and effective $ l\_2 $ norm-based strategy for kv cache compression.
\newblock \emph{arXiv preprint arXiv:2406.11430}, 2024.

\bibitem[Guo et~al.(2024)Guo, Kamigaito, and Watanabe]{guo2024attention}
Zhiyu Guo, Hidetaka Kamigaito, and Taro Watanabe.
\newblock Attention score is not all you need for token importance indicator in kv cache reduction: Value also matters.
\newblock \emph{arXiv preprint arXiv:2406.12335}, 2024.

\bibitem[Long et~al.(2023)Long, Zhao, Pi, Wang, and Wang]{long2023beyond}
Sifan Long, Zhen Zhao, Jimin Pi, Shengsheng Wang, and Jingdong Wang.
\newblock Beyond attentive tokens: Incorporating token importance and diversity for efficient vision transformers.
\newblock In \emph{Proceedings of the IEEE/CVF Conference on Computer Vision and Pattern Recognition}, pages 10334--10343, 2023.

\bibitem[Goemans and Williamson(1995)]{goemans1995improved}
Michel~X Goemans and David~P Williamson.
\newblock Improved approximation algorithms for maximum cut and satisfiability problems using semidefinite programming.
\newblock \emph{Journal of the ACM (JACM)}, 42\penalty0 (6):\penalty0 1115--1145, 1995.

\bibitem[Charikar(2002)]{charikar2002similarity}
Moses~S Charikar.
\newblock Similarity estimation techniques from rounding algorithms.
\newblock In \emph{Proceedings of the thiry-fourth annual ACM symposium on Theory of computing}, pages 380--388, 2002.

\bibitem[Cobbe et~al.(2021)Cobbe, Kosaraju, Bavarian, Chen, Jun, Kaiser, Plappert, Tworek, Hilton, Nakano, et~al.]{cobbe2021training}
Karl Cobbe, Vineet Kosaraju, Mohammad Bavarian, Mark Chen, Heewoo Jun, Lukasz Kaiser, Matthias Plappert, Jerry Tworek, Jacob Hilton, Reiichiro Nakano, et~al.
\newblock Training verifiers to solve math word problems.
\newblock \emph{arXiv preprint arXiv:2110.14168}, 2021.

\bibitem[Hsieh et~al.(2024)Hsieh, Sun, Kriman, Acharya, Rekesh, Jia, and Ginsburg]{hsieh2024ruler}
Cheng-Ping Hsieh, Simeng Sun, Samuel Kriman, Shantanu Acharya, Dima Rekesh, Fei Jia, and Boris Ginsburg.
\newblock Ruler: What's the real context size of your long-context language models?
\newblock \emph{arXiv preprint arXiv:2404.06654}, 2024.

\bibitem[Bai et~al.(2023)Bai, Lv, Zhang, Lyu, Tang, Huang, Du, Liu, Zeng, Hou, et~al.]{bai2023longbench}
Yushi Bai, Xin Lv, Jiajie Zhang, Hongchang Lyu, Jiankai Tang, Zhidian Huang, Zhengxiao Du, Xiao Liu, Aohan Zeng, Lei Hou, et~al.
\newblock Longbench: A bilingual, multitask benchmark for long context understanding.
\newblock \emph{arXiv preprint arXiv:2308.14508}, 2023.

\bibitem[Phuong and Hutter(2022)]{phuong2022formal}
Mary Phuong and Marcus Hutter.
\newblock Formal algorithms for transformers.
\newblock \emph{arXiv preprint arXiv:2207.09238}, 2022.

\bibitem[Andoni et~al.(2018)Andoni, Indyk, and Razenshteyn]{andoni2018approximate}
Alexandr Andoni, Piotr Indyk, and Ilya Razenshteyn.
\newblock Approximate nearest neighbor search in high dimensions.
\newblock In \emph{Proceedings of the International Congress of Mathematicians: Rio de Janeiro 2018}, pages 3287--3318. World Scientific, 2018.

\bibitem[Liu et~al.(2024{\natexlab{a}})Liu, Desai, Liao, Wang, Xie, Xu, Kyrillidis, and Shrivastava]{liu2024scissorhands}
Zichang Liu, Aditya Desai, Fangshuo Liao, Weitao Wang, Victor Xie, Zhaozhuo Xu, Anastasios Kyrillidis, and Anshumali Shrivastava.
\newblock Scissorhands: Exploiting the persistence of importance hypothesis for llm kv cache compression at test time.
\newblock \emph{Advances in Neural Information Processing Systems}, 36, 2024{\natexlab{a}}.

\bibitem[Ainslie et~al.(2023)Ainslie, Lee-Thorp, de~Jong, Zemlyanskiy, Lebr{\'o}n, and Sanghai]{ainslie2023gqa}
Joshua Ainslie, James Lee-Thorp, Michiel de~Jong, Yury Zemlyanskiy, Federico Lebr{\'o}n, and Sumit Sanghai.
\newblock Gqa: Training generalized multi-query transformer models from multi-head checkpoints.
\newblock \emph{arXiv preprint arXiv:2305.13245}, 2023.

\bibitem[Wang et~al.(2024)Wang, Jin, Yu, and Zhang]{wang2024model}
Zheng Wang, Boxiao Jin, Zhongzhi Yu, and Minjia Zhang.
\newblock Model tells you where to merge: Adaptive kv cache merging for llms on long-context tasks.
\newblock \emph{arXiv preprint arXiv:2407.08454}, 2024.

\bibitem[Liu et~al.(2024{\natexlab{b}})Liu, Liu, Pan, He, Haffari, and Zhuang]{liu2024minicache}
Akide Liu, Jing Liu, Zizheng Pan, Yefei He, Gholamreza Haffari, and Bohan Zhuang.
\newblock Minicache: Kv cache compression in depth dimension for large language models.
\newblock \emph{arXiv preprint arXiv:2405.14366}, 2024{\natexlab{b}}.

\bibitem[Kitaev et~al.(2020)Kitaev, Kaiser, and Levskaya]{kitaev2020reformer}
Nikita Kitaev, {\L}ukasz Kaiser, and Anselm Levskaya.
\newblock Reformer: The efficient transformer.
\newblock \emph{arXiv preprint arXiv:2001.04451}, 2020.

\bibitem[Zandieh et~al.(2023)Zandieh, Han, Daliri, and Karbasi]{zandieh2023kdeformer}
Amir Zandieh, Insu Han, Majid Daliri, and Amin Karbasi.
\newblock Kdeformer: Accelerating transformers via kernel density estimation.
\newblock In \emph{International Conference on Machine Learning}, pages 40605--40623. PMLR, 2023.

\bibitem[Han et~al.(2023)Han, Jayaram, Karbasi, Mirrokni, Woodruff, and Zandieh]{han2023hyperattention}
Insu Han, Rajesh Jayaram, Amin Karbasi, Vahab Mirrokni, David~P Woodruff, and Amir Zandieh.
\newblock Hyperattention: Long-context attention in near-linear time.
\newblock \emph{arXiv preprint arXiv:2310.05869}, 2023.

\bibitem[Zandieh et~al.(2024{\natexlab{a}})Zandieh, Daliri, and Han]{zandieh2024qjl}
Amir Zandieh, Majid Daliri, and Insu Han.
\newblock Qjl: 1-bit quantized jl transform for kv cache quantization with zero overhead.
\newblock \emph{arXiv preprint arXiv:2406.03482}, 2024{\natexlab{a}}.

\bibitem[Zandieh et~al.(2024{\natexlab{b}})Zandieh, Han, Mirrokni, and Karbasi]{zandieh2024subgen}
Amir Zandieh, Insu Han, Vahab Mirrokni, and Amin Karbasi.
\newblock Subgen: Token generation in sublinear time and memory.
\newblock \emph{arXiv preprint arXiv:2402.06082}, 2024{\natexlab{b}}.

\bibitem[Shazeer(2019)]{shazeer2019fast}
Noam Shazeer.
\newblock Fast transformer decoding: One write-head is all you need.
\newblock \emph{arXiv preprint arXiv:1911.02150}, 2019.

\bibitem[Hooper et~al.(2024)Hooper, Kim, Mohammadzadeh, Mahoney, Shao, Keutzer, and Gholami]{hooper2024kvquant}
Coleman Hooper, Sehoon Kim, Hiva Mohammadzadeh, Michael~W Mahoney, Yakun~Sophia Shao, Kurt Keutzer, and Amir Gholami.
\newblock Kvquant: Towards 10 million context length llm inference with kv cache quantization.
\newblock \emph{arXiv preprint arXiv:2401.18079}, 2024.

\bibitem[Sheng et~al.(2023)Sheng, Zheng, Yuan, Li, Ryabinin, Chen, Liang, R{\'e}, Stoica, and Zhang]{sheng2023flexgen}
Ying Sheng, Lianmin Zheng, Binhang Yuan, Zhuohan Li, Max Ryabinin, Beidi Chen, Percy Liang, Christopher R{\'e}, Ion Stoica, and Ce~Zhang.
\newblock Flexgen: High-throughput generative inference of large language models with a single gpu.
\newblock In \emph{International Conference on Machine Learning}, pages 31094--31116. PMLR, 2023.

\bibitem[Katharopoulos et~al.(2020)Katharopoulos, Vyas, Pappas, and Fleuret]{katharopoulos2020transformers}
Angelos Katharopoulos, Apoorv Vyas, Nikolaos Pappas, and Fran{\c{c}}ois Fleuret.
\newblock Transformers are rnns: Fast autoregressive transformers with linear attention.
\newblock In \emph{International conference on machine learning}, pages 5156--5165. PMLR, 2020.

\bibitem[Strati et~al.(2024)Strati, Mcallister, Phanishayee, Tarnawski, and Klimovic]{strati2024d}
Foteini Strati, Sara Mcallister, Amar Phanishayee, Jakub Tarnawski, and Ana Klimovic.
\newblock D$\backslash$'ej$\backslash$avu: Kv-cache streaming for fast, fault-tolerant generative llm serving.
\newblock \emph{arXiv preprint arXiv:2403.01876}, 2024.

\bibitem[Jin et~al.(2021)Jin, Pan, Oufattole, Weng, Fang, and Szolovits]{jin2021disease}
Di~Jin, Eileen Pan, Nassim Oufattole, Wei-Hung Weng, Hanyi Fang, and Peter Szolovits.
\newblock What disease does this patient have? a large-scale open domain question answering dataset from medical exams.
\newblock \emph{Applied Sciences}, 11\penalty0 (14):\penalty0 6421, 2021.

\bibitem[Xiao et~al.(2023)Xiao, Tian, Chen, Han, and Lewis]{xiao2023efficient}
Guangxuan Xiao, Yuandong Tian, Beidi Chen, Song Han, and Mike Lewis.
\newblock Efficient streaming language models with attention sinks.
\newblock \emph{arXiv preprint arXiv:2309.17453}, 2023.

\bibitem[Child et~al.(2019)Child, Gray, Radford, and Sutskever]{child2019generating}
Rewon Child, Scott Gray, Alec Radford, and Ilya Sutskever.
\newblock Generating long sequences with sparse transformers.
\newblock \emph{arXiv preprint arXiv:1904.10509}, 2019.

\bibitem[Beltagy et~al.(2020)Beltagy, Peters, and Cohan]{beltagy2020longformer}
Iz~Beltagy, Matthew~E Peters, and Arman Cohan.
\newblock Longformer: The long-document transformer.
\newblock \emph{arXiv preprint arXiv:2004.05150}, 2020.

\bibitem[Gupta et~al.(2021)Gupta, Dar, Goodman, Ciprut, and Berant]{gupta2021memory}
Ankit Gupta, Guy Dar, Shaya Goodman, David Ciprut, and Jonathan Berant.
\newblock Memory-efficient transformers via top-$ k $ attention.
\newblock \emph{arXiv preprint arXiv:2106.06899}, 2021.

\end{thebibliography}

\newpage
\appendix
\section*{Appendix}

\section{Question-Answering Granular Experiment Results}
\label{detail-results}

\begin{table}[htbp]
\centering
\caption{GSM8K and MedQA Question Answering BERTScore}

\begin{tabular}{@{}cc|ccc|ccc@{}}
\toprule
 &  & \multicolumn{3}{c|}{GSM8K} & \multicolumn{3}{c}{Medqa} \\ \midrule
\begin{tabular}[c]{@{}c@{}}Cache Budget / \\ Fastgen Attn \\ Recovery Frac (\%) \end{tabular} & Strategy & Precision & Recall & F1 & Precision & Recall & F1 \\ \midrule
\multirow{4}{*}{10} & \ours & 0.859 & 0.806 & 0.831 & 0.857 & 0.808 & 0.832 \\
 & $L_2$ & 0.858 & 0.798 & 0.826 & 0.833 & 0.813 & 0.823 \\
 & H$_2$O & 0.877 & 0.830 & 0.853 & 0.866 & 0.795 & 0.829 \\
 & Scissorhands & 0.873 & 0.825 & 0.848 & 0.867 & 0.795 & 0.829 \\ \midrule
\multirow{4}{*}{30} & \ours & 0.893 & 0.854 & 0.873 & 0.867 & 0.834 & 0.850 \\
 & $L_2$ & 0.885 & 0.847 & 0.865 & 0.855 & 0.834 & 0.844 \\
 & H$_2$O & 0.893 & 0.860 & 0.877 & 0.878 & 0.802 & 0.838 \\
 & Scissorhands & 0.893 & 0.858 & 0.875 & 0.877 & 0.802 & 0.838 \\ \midrule
\multirow{4}{*}{50} & \ours & 0.897 & 0.865 & 0.880 & 0.869 & 0.842 & 0.855 \\
 & $L_2$ & 0.891 & 0.861 & 0.875 & 0.866 & 0.841 & 0.853 \\
 & H$_2$O & 0.896 & 0.866 & 0.881 & 0.879 & 0.803 & 0.839 \\
 & Scissorhands & 0.896 & 0.864 & 0.879 & 0.878 & 0.804 & 0.839 \\ \midrule
\multirow{4}{*}{70} & \ours & 0.896 & 0.866 & 0.881 & 0.869 & 0.843 & 0.855 \\
 & $L_2$ & 0.894 & 0.865 & 0.879 & 0.868 & 0.842 & 0.855 \\
 & H$_2$O & 0.897 & 0.867 & 0.881 & 0.879 & 0.801 & 0.838 \\
 & Scissorhands & 0.896 & 0.864 & 0.880 & 0.879 & 0.803 & 0.839 \\ \midrule
\multirow{4}{*}{90} & \ours & 0.897 & 0.867 & 0.881 & 0.868 & 0.843 & 0.855 \\
 & $L_2$ & 0.896 & 0.866 & 0.881 & 0.868 & 0.843 & 0.855 \\
 & H$_2$O & 0.897 & 0.867 & 0.881 & 0.879 & 0.801 & 0.838 \\
 & Scissorhands & 0.896 & 0.864 & 0.880 & 0.880 & 0.802 & 0.839 \\ \midrule
50 & \multirow{5}{*}{Fastgen} & 0.811 & 0.770 & 0.789 & 0.816 & 0.763 & 0.788 \\
60 &  & 0.827 & 0.778 & 0.801 & 0.806 & 0.766 & 0.785 \\
70 &  & 0.837 & 0.788 & 0.811 & 0.811 & 0.766 & 0.787 \\
80 &  & 0.874 & 0.840 & 0.857 & 0.866 & 0.793 & 0.828 \\
90 &  & 0.896 & 0.864 & 0.879 & 0.876 & 0.800 & 0.836 \\ \midrule
100 & Full & 0.897 & 0.867 & 0.882 & 0.868 & 0.843 & 0.855 \\ \bottomrule
\end{tabular}

\end{table}

\begin{table}[htbp]
\centering
\caption{GSM8K Question Answering Rouge}

\begin{tabular}{@{}cc|cccc@{}}
\toprule
\begin{tabular}[c]{@{}c@{}}Cache Budget /\\ Fastgen Attn\\ Recovery Fra(\%)\end{tabular} & Strategy & Rouge 1 & Rouge 2 & Rouge L & Rouge Lsum \\ \midrule
\multirow{4}{*}{10} & \ours & 0.206 & 0.051 & 0.157 & 0.186 \\
 & $L_2$ & 0.196 & 0.050 & 0.151 & 0.179 \\
 & H$_2$O & 0.300 & 0.090 & 0.227 & 0.263 \\
 & Scissorhands & 0.271 & 0.074 & 0.205 & 0.238 \\ \midrule
\multirow{4}{*}{30} & \ours & 0.446 & 0.187 & 0.341 & 0.383 \\
 & $L_2$ & 0.392 & 0.149 & 0.288 & 0.337 \\
 & H$_2$O & 0.481 & 0.208 & 0.364 & 0.410 \\
 & Scissorhands & 0.471 & 0.203 & 0.357 & 0.403 \\ \midrule
\multirow{4}{*}{50} & \ours & 0.511 & 0.234 & 0.393 & 0.438 \\
 & $L_2$ & 0.476 & 0.205 & 0.355 & 0.409 \\
 & H$_2$O & 0.517 & 0.238 & 0.398 & 0.442 \\
 & Scissorhands & 0.509 & 0.232 & 0.389 & 0.433 \\ \midrule
\multirow{4}{*}{70} & \ours & 0.521 & 0.240 & 0.401 & 0.446 \\
 & $L_2$ & 0.509 & 0.230 & 0.386 & 0.435 \\
 & H$_2$O & 0.523 & 0.243 & 0.404 & 0.446 \\
 & Scissorhands & 0.510 & 0.233 & 0.392 & 0.435 \\ \midrule
\multirow{4}{*}{90} & \ours & 0.525 & 0.243 & 0.403 & 0.449 \\
 & $L_2$ & 0.522 & 0.241 & 0.400 & 0.446 \\
 & H$_2$O & 0.523 & 0.243 & 0.406 & 0.446 \\
 & Scissorhands & 0.512 & 0.235 & 0.393 & 0.436 \\ \midrule
50 & \multirow{5}{*}{Fastgen} & 0.112 & 0.017 & 0.095 & 0.106 \\
60 &  & 0.133 & 0.024 & 0.113 & 0.126 \\
70 &  & 0.171 & 0.036 & 0.139 & 0.160 \\
80 &  & 0.356 & 0.128 & 0.264 & 0.305 \\
90 &  & 0.509 & 0.231 & 0.391 & 0.434 \\ \midrule
100 & Full & 0.526 & 0.244 & 0.403 & 0.449 \\ \bottomrule
\end{tabular}

\end{table}

\begin{table}[htbp]
\centering
\caption{GSM8K Question Answering GPT4-Judge}

\begin{tabular}{@{}cc|cccc@{}}
\toprule
\begin{tabular}[c]{@{}c@{}}Cache Budget / \\ Fastgen Attn \\ Recovery Frac (\%) \end{tabular} & Strategy & Similarity to GT & Coherence & Faithfulness & Helpfulness \\ \midrule
\multirow{4}{*}{10} & \ours & 1.018 & 1.387 & 1.147 & 1.083 \\
 & $L_2$ & 1.005 & 1.293 & 1.098 & 1.033 \\
 & H$_2$O & 1.172 & 2.304 & 1.566 & 1.630 \\
 & Scissorhands & 1.138 & 2.132 & 1.424 & 1.452 \\ \midrule
\multirow{4}{*}{30} & \ours & 2.520 & 3.767 & 3.216 & 3.190 \\
 & $L_2$ & 1.356 & 2.428 & 1.895 & 1.841 \\
 & H$_2$O & 3.014 & 4.252 & 3.706 & 3.860 \\
 & Scissorhands & 2.906 & 4.184 & 3.636 & 3.798 \\ \midrule
\multirow{4}{*}{50} & \ours & 3.457 & 4.530 & 4.212 & 4.241 \\
 & $L_2$ & 2.190 & 3.494 & 3.035 & 3.027 \\
 & H$_2$O & 3.798 & 4.712 & 4.434 & 4.534 \\
 & Scissorhands & 3.582 & 4.604 & 4.276 & 4.400 \\ \midrule
\multirow{4}{*}{70} & \ours & 3.734 & 4.671 & 4.404 & 4.444 \\
 & $L_2$ & 2.934 & 4.184 & 3.817 & 3.820 \\
 & H$_2$O & 3.940 & 4.774 & 4.576 & 4.656 \\
 & Scissorhands & 3.712 & 4.668 & 4.334 & 4.462 \\ \midrule
\multirow{4}{*}{90} & \ours & 3.569 & 4.578 & 4.324 & 4.361 \\
 & $L_2$ & 3.837 & 4.722 & 4.468 & 4.525 \\
 & H$_2$O & 3.970 & 4.814 & 4.596 & 4.688 \\
 & Scissorhands & 3.750 & 4.676 & 4.392 & 4.504 \\ \midrule
50 & \multirow{5}{*}{Fastgen} & 1.000 & 1.074 & 1.040 & 1.028 \\
60 &  & 1.000 & 1.054 & 1.022 & 1.010 \\
70 &  & 1.008 & 1.116 & 1.048 & 1.014 \\
80 &  & 1.472 & 2.602 & 2.118 & 2.234 \\
90 &  & 3.838 & 4.714 & 4.448 & 4.554 \\ \midrule
100 & Full & 3.845 & 4.716 & 4.499 & 4.545 \\ \bottomrule
\end{tabular}

\end{table}

\begin{table}[htbp]
\centering
\caption{MedQA Question Answering BERTScore}

\begin{tabular}{@{}ccrrr@{}}
\toprule
\begin{tabular}[c]{@{}c@{}}Cache Budget /\\ Fastgen Attn\\ Recovery Fra(\%)\end{tabular} & Strategy & \multicolumn{1}{c}{Precision} & \multicolumn{1}{c}{Recall} & \multicolumn{1}{c}{F1} \\ \midrule
\multirow{4}{*}{10} & \multicolumn{1}{c|}{\ours} & 0.857 & 0.808 & 0.832 \\
 & \multicolumn{1}{c|}{$L_2$} & 0.833 & 0.813 & 0.823 \\
 & \multicolumn{1}{c|}{H$_2$O} & 0.866 & 0.795 & 0.829 \\
 & \multicolumn{1}{c|}{Scissorhands} & 0.867 & 0.795 & 0.829 \\ \midrule
\multirow{4}{*}{30} & \multicolumn{1}{c|}{\ours} & 0.867 & 0.834 & 0.850 \\
 & \multicolumn{1}{c|}{$L_2$} & 0.855 & 0.834 & 0.844 \\
 & \multicolumn{1}{c|}{H$_2$O} & 0.878 & 0.802 & 0.838 \\
 & \multicolumn{1}{c|}{Scissorhands} & 0.877 & 0.802 & 0.838 \\ \midrule
\multirow{4}{*}{50} & \multicolumn{1}{c|}{\ours} & 0.869 & 0.842 & 0.855 \\
 & \multicolumn{1}{c|}{$L_2$} & 0.866 & 0.841 & 0.853 \\
 & \multicolumn{1}{c|}{H$_2$O} & 0.879 & 0.803 & 0.839 \\
 & \multicolumn{1}{c|}{Scissorhands} & 0.878 & 0.804 & 0.839 \\ \midrule
\multirow{4}{*}{70} & \multicolumn{1}{c|}{\ours} & 0.869 & 0.843 & 0.855 \\
 & \multicolumn{1}{c|}{$L_2$} & 0.868 & 0.842 & 0.855 \\
 & \multicolumn{1}{c|}{H$_2$O} & 0.879 & 0.801 & 0.838 \\
 & \multicolumn{1}{c|}{Scissorhands} & 0.879 & 0.803 & 0.839 \\ \midrule
\multirow{4}{*}{90} & \multicolumn{1}{c|}{\ours} & 0.868 & 0.843 & 0.855 \\
 & \multicolumn{1}{c|}{$L_2$} & 0.868 & 0.843 & 0.855 \\
 & \multicolumn{1}{c|}{H$_2$O} & 0.879 & 0.801 & 0.838 \\
 & \multicolumn{1}{c|}{Scissorhands} & 0.880 & 0.802 & 0.839 \\ \midrule
50 & \multicolumn{1}{c|}{\multirow{5}{*}{Fastgen}} & 0.816 & 0.763 & 0.788 \\
60 & \multicolumn{1}{c|}{} & 0.806 & 0.766 & 0.785 \\
70 & \multicolumn{1}{c|}{} & 0.811 & 0.766 & 0.787 \\
80 & \multicolumn{1}{c|}{} & 0.866 & 0.793 & 0.828 \\
90 & \multicolumn{1}{c|}{} & 0.876 & 0.800 & 0.836 \\ \midrule
100 & Full & 0.868 & 0.843 & 0.855 \\ \bottomrule
\end{tabular}

\end{table}
\begin{table}[htbp]
\centering
\caption{MedQA Question Answering Rouge}

\begin{tabular}{@{}cccccc@{}}
\toprule
Cache Budget (\%) & Strategy & Rouge 1 & Rouge 2 & Rouge L & Rouge Lsum \\ \midrule
\multirow{4}{*}{10} & \multicolumn{1}{c|}{\ours} & 0.346 & 0.110 & 0.171 & 0.324 \\
 & \multicolumn{1}{c|}{$L_2$} & 0.304 & 0.072 & 0.154 & 0.289 \\
 & \multicolumn{1}{c|}{H$_2$O} & 0.236 & 0.092 & 0.138 & 0.220 \\
 & \multicolumn{1}{c|}{Scissorhands} & 0.237 & 0.091 & 0.139 & 0.221 \\ \midrule
\multirow{4}{*}{30} & \multicolumn{1}{c|}{\ours} & 0.449 & 0.170 & 0.227 & 0.426 \\
 & \multicolumn{1}{c|}{$L_2$} & 0.429 & 0.146 & 0.213 & 0.407 \\
 & \multicolumn{1}{c|}{H$_2$O} & 0.255 & 0.118 & 0.151 & 0.239 \\
 & \multicolumn{1}{c|}{Scissorhands} & 0.252 & 0.116 & 0.151 & 0.236 \\ \midrule
\multirow{4}{*}{50} & \multicolumn{1}{c|}{\ours} & 0.481 & 0.194 & 0.245 & 0.455 \\
 & \multicolumn{1}{c|}{$L_2$} & 0.474 & 0.184 & 0.240 & 0.449 \\
 & \multicolumn{1}{c|}{H$_2$O} & 0.243 & 0.107 & 0.149 & 0.229 \\
 & \multicolumn{1}{c|}{Scissorhands} & 0.244 & 0.110 & 0.150 & 0.230 \\ \midrule
\multirow{4}{*}{70} & \multicolumn{1}{c|}{\ours} & 0.487 & 0.197 & 0.249 & 0.461 \\
 & \multicolumn{1}{c|}{$L_2$} & 0.484 & 0.194 & 0.247 & 0.458 \\
 & \multicolumn{1}{c|}{H$_2$O} & 0.229 & 0.097 & 0.143 & 0.216 \\
 & \multicolumn{1}{c|}{Scissorhands} & 0.234 & 0.103 & 0.147 & 0.219 \\ \midrule
\multirow{4}{*}{90} & \multicolumn{1}{c|}{\ours} & 0.487 & 0.197 & 0.249 & 0.461 \\
 & \multicolumn{1}{c|}{$L_2$} & 0.487 & 0.197 & 0.249 & 0.461 \\
 & \multicolumn{1}{c|}{H$_2$O} & 0.223 & 0.095 & 0.142 & 0.211 \\
 & \multicolumn{1}{c|}{Scissorhands} & 0.228 & 0.099 & 0.145 & 0.214 \\ \midrule
50 & \multicolumn{1}{c|}{\multirow{5}{*}{Fastgen}} & 0.068 & 0.013 & 0.052 & 0.066 \\
60 & \multicolumn{1}{c|}{} & 0.079 & 0.014 & 0.061 & 0.077 \\
70 & \multicolumn{1}{c|}{} & 0.103 & 0.020 & 0.074 & 0.099 \\
80 & \multicolumn{1}{c|}{} & 0.208 & 0.069 & 0.126 & 0.192 \\
90 & \multicolumn{1}{c|}{} & 0.220 & 0.092 & 0.140 & 0.207 \\ \midrule
100 & Full & 0.486 & 0.198 & 0.248 & 0.460 \\ \bottomrule
\end{tabular}

\end{table}
\begin{table}[htbp]
\centering
\caption{MedQA Question Answering GPT4-Judge}

\begin{tabular}{@{}cc|cccc@{}}
\toprule
\begin{tabular}[c]{@{}c@{}}Cache Budget /\\ Fastgen Attn\\ Recovery Frac (\%)\end{tabular} & Strategy & Similarity to GT & Coherence & Faithfulness & Helpfulness \\ \midrule
\multirow{4}{*}{10} & \ours & 1.970 & 3.517 & 2.665 & 2.547 \\
 & $L_2$ & 1.103 & 1.695 & 1.639 & 1.283 \\
 & H$_2$O & 2.138 & 3.206 & 2.594 & 2.416 \\
 & Scissorhands & 2.144 & 3.202 & 2.580 & 2.402 \\ \midrule
\multirow{4}{*}{30} & \ours & 2.511 & 4.415 & 3.533 & 3.613 \\
 & $L_2$ & 1.939 & 3.633 & 2.942 & 2.843 \\
 & H$_2$O & 3.428 & 3.818 & 3.608 & 3.276 \\
 & Scissorhands & 3.406 & 3.850 & 3.602 & 3.286 \\ \midrule
\multirow{4}{*}{50} & \ours & 3.022 & 4.730 & 4.139 & 4.254 \\
 & $L_2$ & 2.850 & 4.511 & 3.797 & 3.950 \\
 & H$_2$O & 2.938 & 3.632 & 3.280 & 2.762 \\
 & Scissorhands & 2.918 & 3.634 & 3.308 & 2.748 \\ \midrule
\multirow{4}{*}{70} & \ours & 3.232 & 4.809 & 4.292 & 4.434 \\
 & $L_2$ & 3.194 & 4.755 & 4.235 & 4.385 \\
 & H$_2$O & 2.414 & 3.396 & 2.958 & 2.178 \\
 & Scissorhands & 2.554 & 3.454 & 3.098 & 2.328 \\ \midrule
\multirow{4}{*}{90} & \ours & 3.291 & 4.839 & 4.355 & 4.507 \\
 & $L_2$ & 3.265 & 4.818 & 4.318 & 4.458 \\
 & H$_2$O & 2.400 & 3.232 & 2.830 & 2.016 \\
 & Scissorhands & 2.404 & 3.346 & 2.980 & 2.098 \\ \midrule
50 & \multirow{5}{*}{Fastgen} & 1.002 & 1.004 & 1.006 & 1.000 \\
60 &  & 1.005 & 1.004 & 1.005 & 1.000 \\
70 &  & 1.008 & 1.014 & 1.014 & 1.008 \\
80 &  & 1.620 & 2.783 & 2.270 & 1.512 \\
90 &  & 2.356 & 3.242 & 2.748 & 1.870 \\ \midrule
100 & Full & 3.337 & 4.817 & 4.342 & 4.500 \\ \bottomrule
\end{tabular}

\label{MedQA-qa}
\end{table}

\section{Results of ablation on Hash Dimension}
\label{appendix:lsh-dim-ablation}
Please see Table \ref{table:ablation}
\begin{table}[htbp]
\centering
\caption{\textbf{LSH Hash Dimension Ablation.} We assesses GSM8K Question Answering performance for different LSH dimensions. The cache budget is fixed at 50\%. LSH dimension does not significantly impact performance. Small LSH dimensions slightly outperform larger LSH dimensions.}
\begin{tabular}{@{}c|ccccc@{}}
\toprule
\multicolumn{1}{c|}{\begin{tabular}[c]{@{}c@{}}LSH\\ Dim\end{tabular}} & \multicolumn{1}{c}{\begin{tabular}[c]{@{}c@{}}BERTScore\\F1\end{tabular}} & \multicolumn{1}{c}{Rouge L} & \multicolumn{1}{c}{\begin{tabular}[c]{@{}c@{}}GPT4\\Judge\end{tabular}} & \multicolumn{1}{l}{\begin{tabular}[c]{@{}c@{}}Compression\\ Ratio\end{tabular}} & \multicolumn{1}{c}{\begin{tabular}[c]{@{}c@{}}Cache\\ Memory\\ (GB)\end{tabular}} \\ \midrule
4 & 0.8807 & 0.3974 & 4.3833 & 0.3728 & 2.8062 \\ \midrule
8 & 0.8802 & \textbf{0.3975} & \textbf{4.4113} & 0.3734 & 2.8355 \\ \midrule
16 & \textbf{0.8807} & 0.3972 & 4.3753 & 0.3716 & 2.8941 \\ \midrule
24 & 0.8802 & 0.3951 & 4.3733 & 0.3711 & 2.9527 \\ \midrule
32 & 0.8796 & 0.3926 & 4.3220 & 0.3710 & 3.0113 \\ \midrule
64 & 0.8797 & 0.3900 & 4.2333 & 0.3702 & 3.2456 \\ \bottomrule
\end{tabular}
\label{table:ablation}
\end{table}

\section{Attention Scores and Key Norms Visualization}

\begin{figure}[htbp]
    \centering
    \includegraphics[width=1\linewidth]{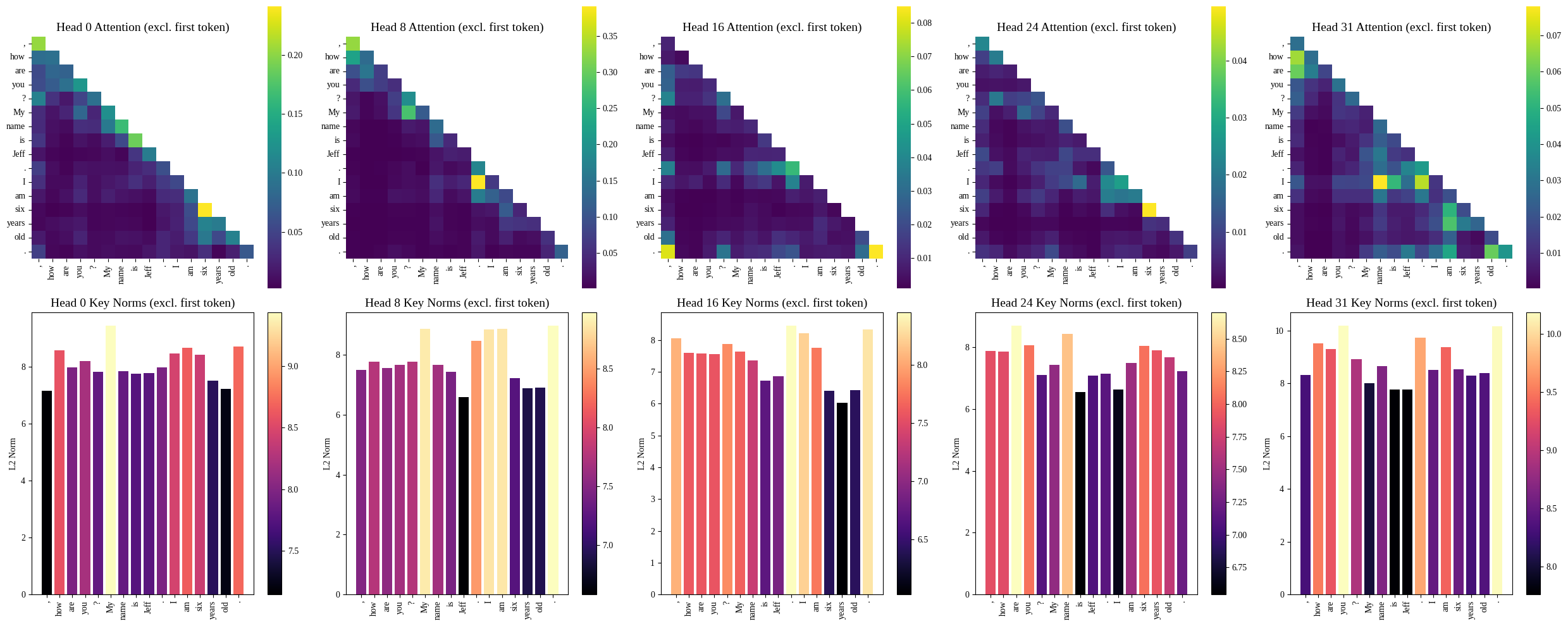}
    \caption{\textbf{Attention and Key Norms.} Attention scores and corresponding $L_2$ norms of key vectors (excluding the first token) for a sample of heads (0,8,16,24,31) in the 8th layer for a sample input sequence. Each subplot shows the attention heatmap (top) and the corresponding key norm values (bottom) for a particular head, allowing for a direct comparison between attention patterns and key norm values across different heads.}
    \label{fig:attn-key-norm}
\end{figure}

We further examine the method of our chief competitor, the $L_2$ eviction method \citep{devoto2024simple}. In particular, in Figure \ref{fig:attn-key-norm} we examine the key-norm-attention correlation suggested by the authors. Indeed, low key-norms, even across prompts, demonstrate a strong correlation with attention score.

\section{Attention Loss Ratio Analysis}
\label{alr-analysis-section}
We perform an attention loss ratio (ALR) analysis between LSH-based ranking and $L_2$-based ranking. Our implementation is an adaptation of the methodology described in \cite{devoto2024simple}. This section explores how much of the uncompressed attention matrix is preserved between \ours and the $L_2$ eviction strategy in \cite{devoto2024simple}. 

Compressing the KV cache entails dropping KV pairs. Per \citep{devoto2024simple}, we can define the attention loss caused by the compression as the sum of the attention scores associated with the dropped KV pairs in layer $l$ and head $h$ 
via the equation $L_{l,h}^m = \sum_{p \in D_{l,h}^m} a_{l,h,p}$, where $a_{l,h,p}$ is the average attention score at position $p$ for layer $l$ and head $h$, and $D_{l,h}^m$ denotes the positions of the $m$ dropped KV pairs, with $|D_{l,h}^m| = m$. We process a selection of prompts and examine how proposed evictions by the $L_2$ eviction strategy and $\ours$ would affect the sum of attention scores.

\begin{figure}[ht]
    \centering
    \begin{subfigure}[t]{0.40\textwidth} 
        \centering
        \includegraphics[width=0.85\linewidth]{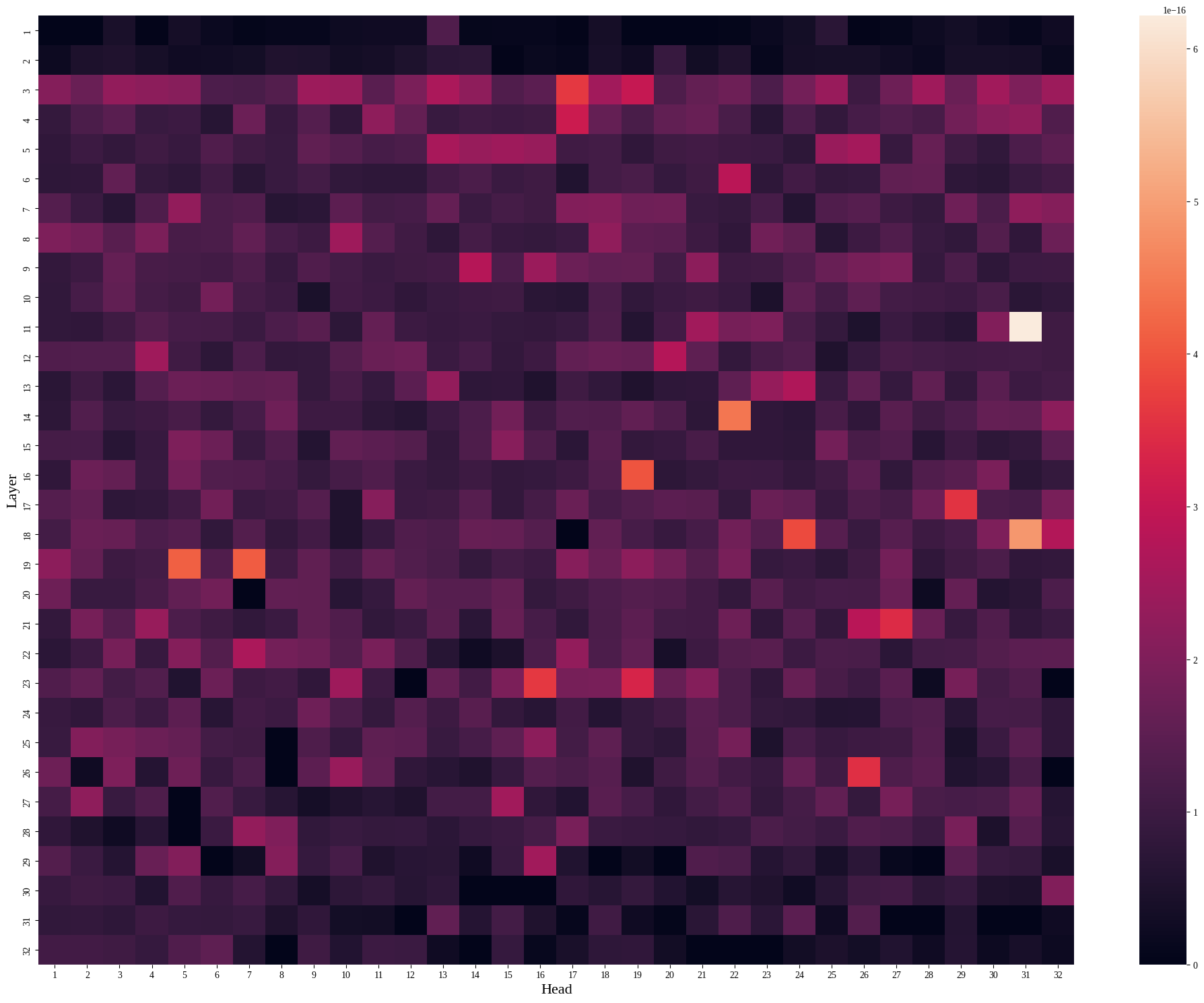} 
        \caption{ALR using LSH ranking}
        \label{fig:alr-lsh}
    \end{subfigure}\hspace{0.02\textwidth}
    \begin{subfigure}[t]{0.40\textwidth} 
        \centering
        \includegraphics[width=0.85\linewidth]{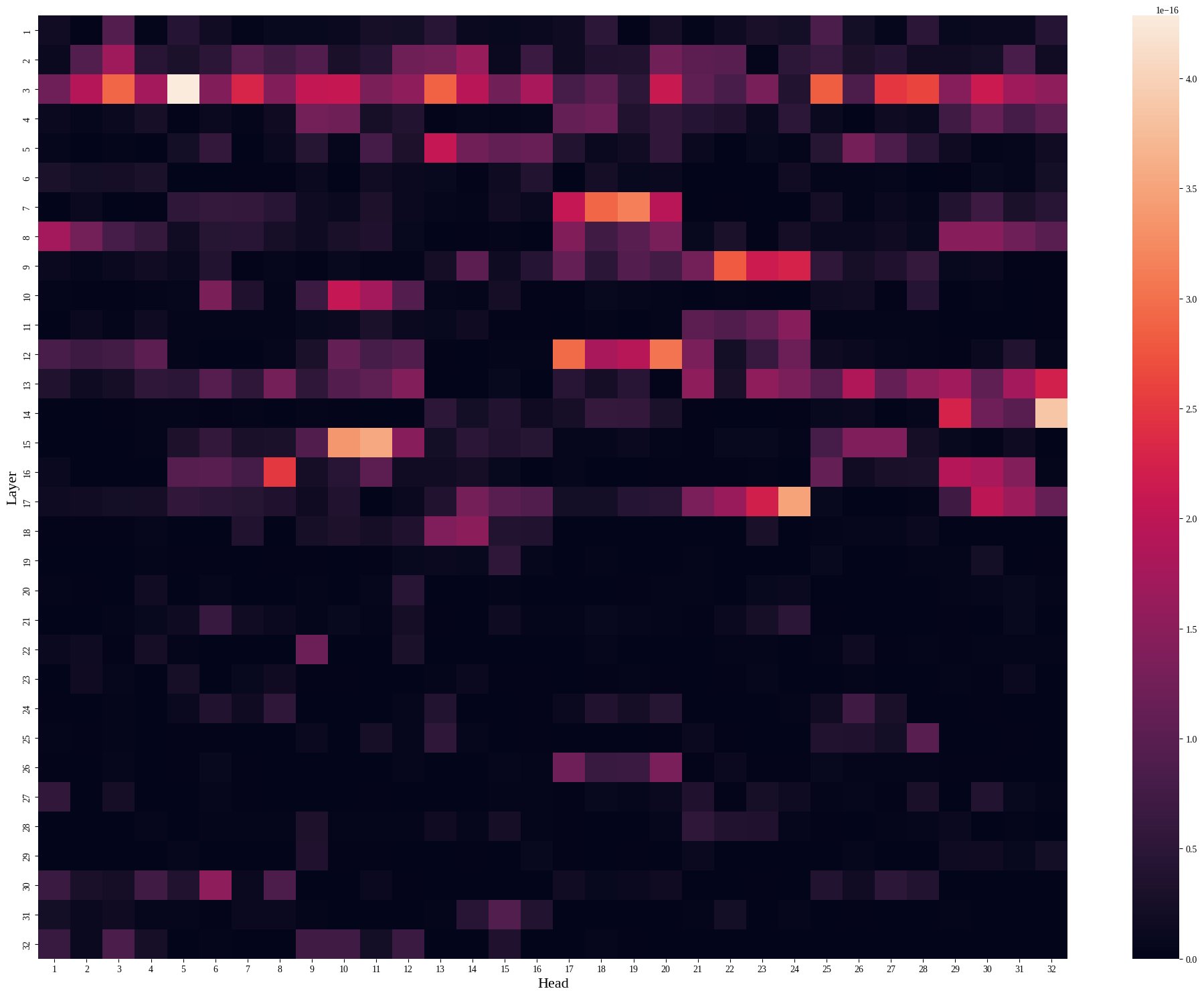} 
        \caption{ALR using $L_2$ ranking}
        \label{fig:alr-l2}
    \end{subfigure}
    \caption{\textbf{Attention Loss Ratio (ALR).} We compare how the eviction strategy of \ours and the $L_2$ method \citep{devoto2024simple} affects the ALR per equation \ref{eq:alr_total}. Our tested model is Llama3-8B-Instruct, which contains 32 heads and 32 attention layers. Cell $(i,j)$ depicts the ALR of head $i$ in attention layer $j$. A darker score indicates a lower ALR. The $L_2$ method exhibits extremely low ALR, thus indicating exclusive preference for high-attention tokens. \ours prefers to select medium-to-high attention tokens.}
    \label{fig:alr}
\end{figure}

To quantify the additional attention loss introduced by using an alternative ranking method (such as $L_2$ norm or \ours's $F_{score}$) instead of the true attention-based ranking, we define the cumulative attention loss difference as:
\begin{equation}
Y_{l,h} = \sum_{m=1}^n \left( L_{l,h}^m - L_{l,h,\text{ref}}^m \right),
\label{eq:alr_total}
\end{equation}
where $L_{l,h,\text{ref}}^m$ is the cumulative attention loss when dropping the KV pairs with the actual lowest attention scores. The value $Y_{l,h}$ is non-negative, and a lower value indicates that the ranking method closely approximates non-compressed attention. Figure \ref{fig:alr} depicts the ALR for the $L_2$ eviction rankings and an LSH ranking.

It is important to note that \ours is not designed to produce a global ranking among the keys as the $L_2$ method is designed to do (via a low-to-high ordering of all $L_2$ key norms). \ours ranks the importance of past tokens with regards to the current token -- and this ranking changes every step. To simulate a comparison, we record the average Hamming distance between the key code of token $i$ and the query codes of all tokens $j>i$. We then sort tokens from lowest to highest average Hamming distance. Figure \ref{fig:alr-lsh} reflects the ALR according to this ranking system. The $L_2$ ranking exclusively prefers high-attention tokens, while the LSH ranking prefers medium-to-high-attention tokens. Based on our empirical results in Section \ref{exps}, the selection of tokens over a spectrum of attention scores skewing towards high results in greater task versatility compared to the $L_2$ eviction.  

\section{On the Analysis of the Relationship between Attention Scores and Average, Accumulated LSH Hamming Distance}

In this section, we follow up on our ALR in Appendix Section \ref{alr-analysis-section}. We analyze the relationship between attention scores and average LSH Hamming distances using 50 randomly selected prompts from GSM8K. We stress that this metric does not perfectly capture the "ranking" system of \ours (which cannot perform a global/full-sequence token-importance ranking like $L_2$ eviction). 

For each prompt, we performed the following:
\begin{enumerate}
    \item \textbf{Captured States}: Extracted normalized key and query vectors from every layer and head combination after applying rotary positional embeddings. 
    \item \textbf{Applied Random Projections}: Applied multiple random Gaussian projections, varying the projection length (number of bits). We tested with projection lengths of 8, 16, 24, and 32.
    \item \textbf{Computed Hamming Distances}: Computed the Hamming distances between the projected and binarized vectors and averaged this over multiple projections to mitigate the randomness that LSH introduces and to obtain a more stable estimate of the Hamming distances.
    \item \textbf{Computed Correlations}: Calculated the Pearson correlation coefficient between the attention scores and the inverted average Hamming distance for each layer and head combination and for each projection length.    
\end{enumerate}

\subsection{Results}
\begin{figure}[ht]
\begin{subfigure}[t]{0.49\textwidth}
    \centering
    \includegraphics[width=1\linewidth]{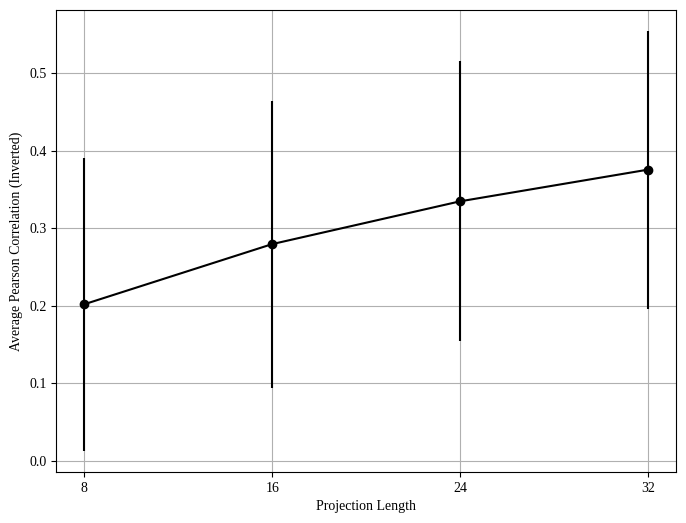}
    \caption{\textbf{Correlations for varying LSH dimension.} We study the Pearson correlations between attention scores and the inverted average Hamming distances, computed over 50 randomly selected prompts from GSM8K, as a function of projection length for Llama-3-8B-Instruct. The tested projection lengths are 8,16, 24, and 32. The error bars indicate the standard deviation. Correlation strengthens as projection length increases.}
    \label{fig:cor vs projection l}
\end{subfigure}\hspace{1em}%
\begin{subfigure}[t]{0.49\textwidth}
    \centering
    \includegraphics[width=1\linewidth]{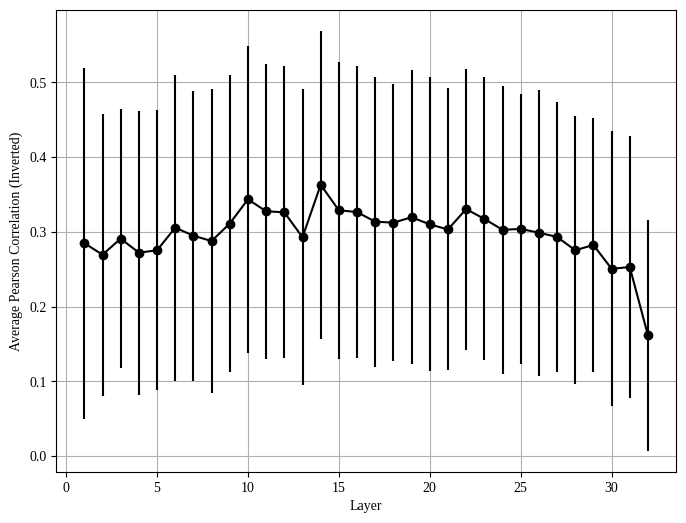}
    \caption{\textbf{Correlations by layer.} We measure the Pearson correlations between attention scores and the inverted average Hamming distances for each transformer layer in Llama-3-8B-Instruct computed over 50 randomly selected prompts from GSM8K. Error bars indicate standard deviation. The final three layers have the weakest correlations.}
    \label{fig:avg_corr_vs_layer}
\end{subfigure}
\begin{subfigure}[t]{0.49\textwidth}
    \centering
    \includegraphics[width=1\linewidth]{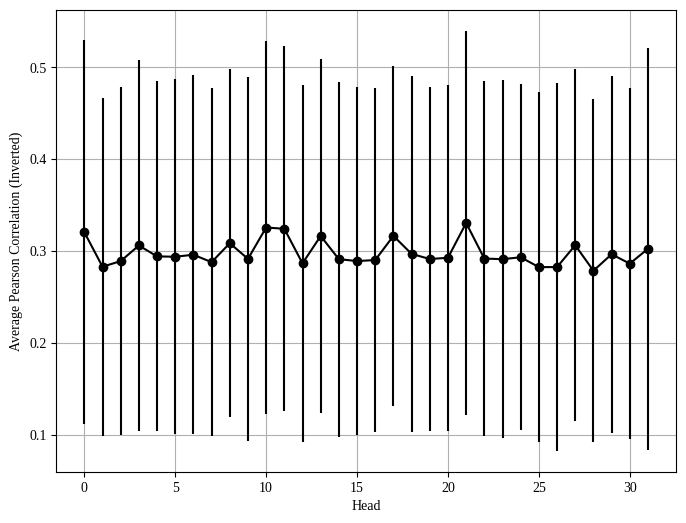}
    \caption{\textbf{Correlations by head.} We study the Pearson correlation between attention scores and the inverted average Hamming distances for each head in Llama-3-8B-Instruct computed over 50 randomly selected prompts from GSM8K. Error bars indicate standard deviation. There is minimal variation between heads.}
    \label{fig:vs-head}
\end{subfigure}\hspace{1em}%
\begin{subfigure}[t]{0.49\textwidth}
    \centering
    \includegraphics[width=1\linewidth]{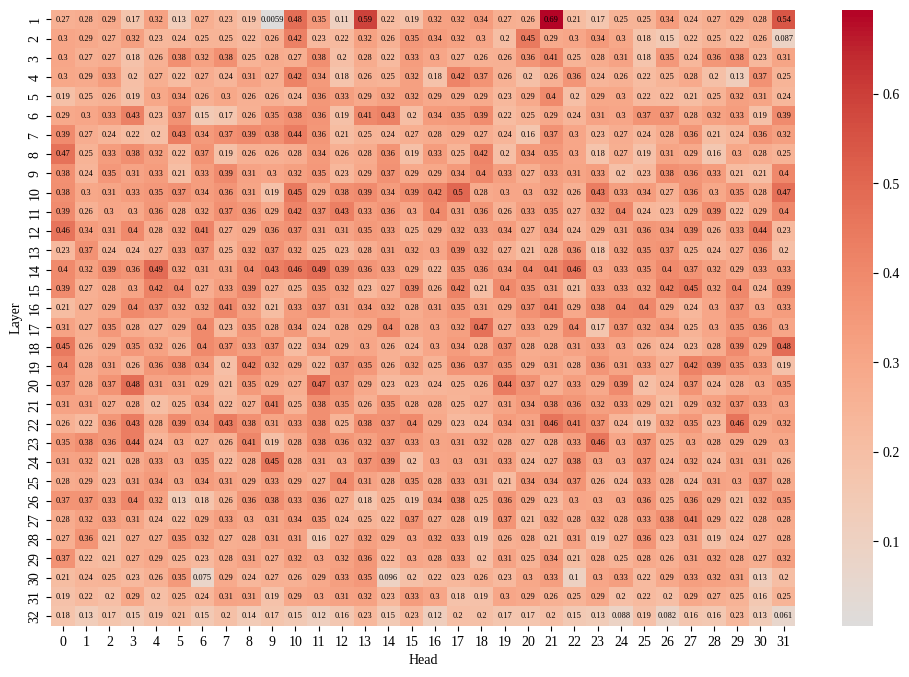}
    \caption{\textbf{Correlation Heat Map.} We examine the average Pearson correlation between attention score and the inverted average Hamming distances (LSH ranking) across all layers and attention heads of Llama-3-8B-Instruct. As attention mass tends to concentrate over a few tokens \citep{gupta2021memory,sheng2023flexgen}, the slightly-weak, but positive correlation indicates the LSH ranking is selecting medium-to-high-attention tokens.}
    \label{fig:pearson}
\end{subfigure}
\caption{\textbf{Correlations of Attention and Inverted Hamming Distances}}
\end{figure}

The average Pearson correlation between the attention scores and the inverted average Hamming distances is $0.2978 \pm 0.1947$.
Table~\ref{tab:avg_corr_proj_length} and Figure~\ref{fig:cor vs projection l} detail the average Pearson correlation per projection length.

\begin{table}[htbp]
\centering
\caption{Average Pearson correlation between attention scores and inverted average Hamming distances per projection length, computed for 50 randomly selected prompts from GSM8k. Higher projection lengths have stronger correlations.}
\begin{tabular}{ccc}
\toprule
Projection Length & Mean & Standard Deviation \\ 
\midrule
8 & 0.2017 & 0.1890 \\
16 & 0.2793 & 0.1852 \\
24 & 0.3345 & 0.1806 \\
32 & 0.3754 & 0.1792 \\
\bottomrule
\end{tabular}
\label{tab:avg_corr_proj_length}
\end{table}

\subsection{Observations}
\begin{itemize}
    \item \textbf{Correlation with Projection Length:} As shown in Figure ~\ref{fig:cor vs projection l} and Table~\ref{tab:avg_corr_proj_length} the average Pearson correlation increases with projection length.  This is likely due to the more detailed vector representation in the projected space, allowing for finer-grained similarity comparisons.
    \item \textbf{Layer-wise Trends:} Figure ~\ref{fig:avg_corr_vs_layer} shows a slight decrease in the average Pearson correlation for the later transformer layers. Earlier layers may be more focused on recognizing broader patterns where the similarity LSH captures is more pronounced compared to the latter layers, which may focus on specifics not captured as effectively by Hamming distances. 
    \item \textbf{Head-wise Consistency:}  The correlation between attention scores and inverted average Hamming distance is relatively consistent across different attention heads, with little variance as seen in Figure `\ref{fig:vs-head}. This uniform behavior indicates that the relationship between attention scores and LSH-measured similarity is, to a large extent, independent of specific head functions.
    \item \textbf{LSH vs. $L_2$ Norms:} While $L_2$ norms were more effective at identifying high-attention tokes, LSH excelled at identifying tokens with moderate attention scores that are vital for the generation of coherent language output. This aligns with the findings of \cite{guo2024attention}, which suggests that tokens with low to medium attention scores are crucial for high-quality language generation. 
    \item \textbf{LSH and Token Similarity:} LSH tended to group tokens together that are similar across dimensions, producing lower Hamming distances. Tokens with very high attention scores may only have strong associations for a relatively small subset of dimensions, which may not always be captured effectively by LSH. 
\end{itemize}

\subsection{ALR Computation Methodology}

We compute the Attention Loss Ratio (ALR) for each layer $l$ and head $h$ as follows:

\begin{enumerate}
    \item 
    \textbf{Data Capture} During the model's forward pass, we capture the necessary data for analysis:
    \begin{itemize}
        \item \textbf{Attention Probabilities} $a_{l,h} \in \mathbb{R}^{n \times n}$: The attention scores between queries and keys.
        \item \textbf{Key Norms} $\|\mathbf{k}_{l,h,p}\|_2$: The $L_2$ norms of key vectors at each position $p$.
        \item \textbf{Key and Query Vectors} $\mathbf{k}_{l,h,p} \in \mathbb{R}^{d}$ and $\mathbf{q}_{l,h,p} \in \mathbb{R}^{d}$: Used for LSH ranking.
    \end{itemize}
    
    \item \textbf{Mean Attention Scores} For each token position $p$, we compute the mean attention score across all positions it attends to:
    \begin{equation}
    \bar{a}_{l,h,p} = \frac{1}{n} \sum_{q=1}^n a_{l,h,p,q}.
    \end{equation}

    \item \textbf{Ranking Methods}
    \begin{itemize}
        \item \textbf{Ideal Attention-Based Ranking} Rank positions in ascending order of $\bar{a}_{l,h,p}$ (from lowest to highest attention score).
        \item \textbf{$L_2$ Norm Ranking} Rank positions in descending order of the key norms $\|\mathbf{k}_{l,h,p}\|_2$.
        \item \textbf{LSH Ranking} Apply Locality-Sensitive Hashing (LSH) to key and query vectors using random projections, compute Hamming distances, and rank positions in ascending order of the average Hamming distance.
    \end{itemize}
    
    \item \textbf{ALR Calculation} For each $m$ from $1$ to $n$, compute the cumulative attention losses:
This allows us to quantitatively compare how well different ranking methods (e.g., $L_2$ norm and LSH ranking) approximate the ideal scenario where the least important KV pairs (those with the lowest attention scores) are dropped during cache compression.

       \begin{align}
   L_{l,h}^m &= \sum_{i=1}^m \bar{a}_{l,h,\pi(i)}, \\
   L_{l,h,\text{ref}}^m &= \sum_{i=1}^m \bar{a}_{l,h,\sigma(i)},
   \end{align}

   where $\pi(i)$ and $\sigma(i)$ are the indices of the $i$-th position in the ranking method and the ideal attention-based ranking, respectively. The ALR for each head and layer is then calculated as $Y_{l,h} = \sum_{m=1}^n \left( L_{l,h}^m - L_{l,h,\text{ref}}^m \right).$

   A lower $Y_{l,h}$ indicates that the ranking method closely approximates the ideal attention-based compression.

\item \textbf{Aggregation} We repeat the above steps for multiple prompts and average the ALR values to obtain the final ALR matrix across layers and heads.
\end{enumerate}

\section{Metrics and Prompts}

\subsection{String Match Score}
\label{appendix:string-match-score}
The string matching score is calculated as: \[
\text{String Matching Score} = \frac{\text{Number of correctly matched characters in predicted string}}{\text{Total number of characters in GT}} \times 100
\]

\subsection{GPT-4-Judge Prompt}
\label{appendix:prompt}
For the GPT-4-Judge metric used in evaluating free response question answering tasks, we accessed the GPT-4o model through OpenAI's API. 

For the GPT4-Rouge metric, the prompt given to the model is:
\begin{lstlisting}
You are shown ground-truth answer(s) and asked to judge the quality of an LLM-generated answer.
Assign it a score from 1-5 where 1 is the worst and 5 is the best based on how similar it is to the ground truth(s).
Do NOT explain your choice. Simply return a number from 1-5.

====GROUND TRUTHS====
{labels}

====ANSWER====
{prediction}
\end{lstlisting}

For the other three GPT4-Judge based on criteria, the prompt given to the model is:
\begin{lstlisting}
You are shown a prompt and asked to assess the quality of an LLM-generated answer on the following dimensions:

===CRITERIA===
{criteria}

Respond with "criteria: score" for each criterion with a newline for each criterion.
Assign a score from 1-5 where 1 is the worst and 5 is the best based on how well the answer meets the criteria.

====PROMPT====
{prompt}

====ANSWER====
{prediction}
\end{lstlisting}

The list of criteria is:

\begin{lstlisting}
CRITERIA = {
    "helpful": "The answer executes the action requested by the prompt without extraneous detail.",
    "coherent": "The answer is logically structured and coherent (ignore the prompt).",
    "faithful": "The answer is faithful to the prompt and does not contain false information.",
}  
\end{lstlisting}

\end{document}